\documentclass[letterpaper, 10 pt, conference]{ieeeconf}  

\IEEEoverridecommandlockouts                              
\usepackage{cite}
\usepackage{amsmath,amssymb,amsfonts}
\usepackage{bbm}
\usepackage{algorithmic}
\usepackage{graphicx}
\usepackage{textcomp}
\usepackage[table]{xcolor}
\usepackage{caption}
\usepackage{multirow}
\usepackage{booktabs}
\usepackage{array}
\usepackage{adjustbox}
\usepackage{url}
\usepackage{hyperref}

\definecolor{gapred}{HTML}{B4533C}
\definecolor{structblue}{HTML}{1F2A44}

\newcolumntype{L}[1]{>{\raggedright\arraybackslash}p{#1}}
\newcolumntype{C}[1]{>{\centering\arraybackslash}p{#1}}

\hypersetup{pdfauthor={}, pdftitle={}, pdfsubject={}, pdfkeywords={}}
\newcommand{\ex}[1]{\textit{#1}}
\newcommand{\vary}[1]{\textcolor{blue!65!black}{\textbf{#1}}}
\title{\LARGE \bf
{The Human--VLA Communication Gap: What Humans Leave Implicit and What Policies Cannot Recover}
}

\author{Yiqing Xu$^{1}$ and Jiajun Wu$^{1}$
\thanks{$^{1}$Yiqing Xu and Jiajun Wu are with the Department of Computer Science, Stanford University, Stanford, CA, USA.
        {\tt\small yiqingx@stanford.edu}}%
}

\begin{document}

\maketitle
\thispagestyle{empty}
\pagestyle{empty}

\begin{abstract}
A robot may know how to perform a task yet fail when a human asks for it naturally. We study this human--VLA communication gap by separating an \emph{information gap}, where task semantics, grounding, or event structure remain unresolved, from a residual \emph{execution gap}, where even a fully specified task depends on how it is organized and invoked. To do this, we characterize how people naturally express demonstrated objectives and evaluate controlled variants of the same physical tasks, with detailed analysis on $\pi_{0.5}$ and consistent trends on $\pi_0$-FAST and Cosmos3-Edge. Our study reveals that humans increasingly compress instructions as tasks grow, relying more on contextual grounding and implicit execution structure. Current policies are misaligned with this behavior: they tolerate grounding dependence relatively well, but struggle when task semantics or event structure must be reconstructed. Even after the information gap is resolved, execution remains sensitive to command granularity and invocation state. These findings identify concrete targets for human-centered policy learning and interfaces that better bridge natural human objectives to robot behavior.
Project website: \url{https://commgap.github.io/}.
\end{abstract}


\section{Introduction}

\begin{figure*}[t]
    \centering
    \includegraphics[width=1\textwidth]{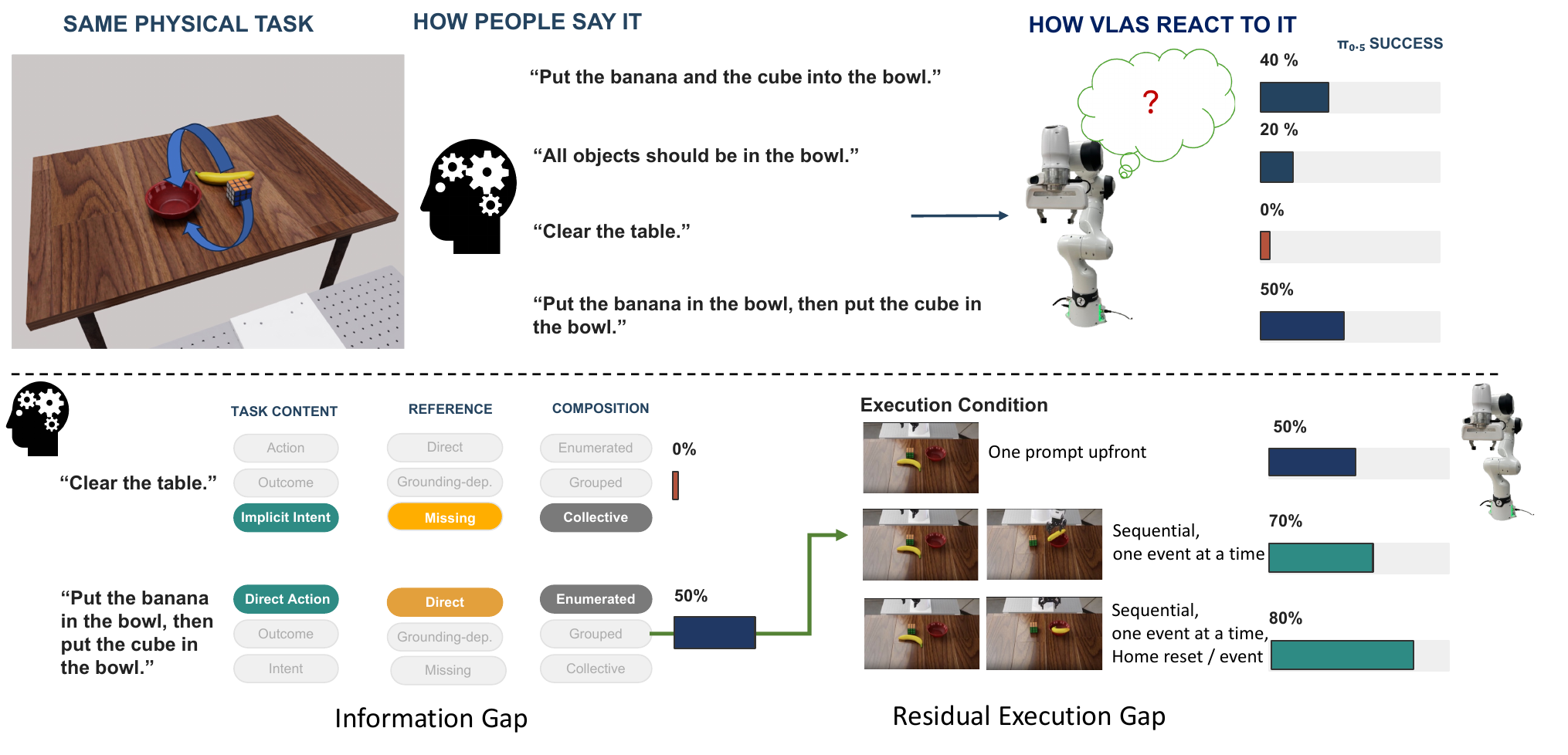}
    \caption{\textbf{Human--VLA communication gap.}
For the same physical task, human specifications vary in task content, reference, and composition, producing substantially different VLA performance (top). We separate this mismatch into an \emph{information gap}, where task semantics, grounding, or event structure remain unresolved, and a residual \emph{execution gap}, where the fully resolved task remains sensitive to command organization and invocation state (bottom), distinguishing recovering the intended task from reliably executing it through the policy's learned interface.}
    \label{fig:communication_gap}
    \vspace{-10pt}
\end{figure*}

Imagine asking a household robot, ``Can you clear the table so I can work here?'' The robot has already learned how to pick up cups, move plates, and place objects into containers, yet it freezes. Replace the request with ``Pick up the white mug and place it in the bin,'' and the same policy succeeds. What failed was not the physical skill but the interface to it: the human expressed an objective naturally, while the policy expected a narrow specification of what to do, on which object, and toward what target state.

This gap matters more as vision--language--action (VLA) models become general-purpose. Recent VLAs acquire diverse manipulation behaviors from large robot datasets and are evaluated by the breadth of tasks, objects, and environments they handle, so existing work has focused on the \emph{robot side} of generality: how many skills are acquired, how they compose, and how they transfer. Yet broad physical competence alone does not make a policy accessible to ordinary users, who should not need to know the command templates used during training. Humans communicate physical objectives by saying an action, describing the outcome, or making an indirect request, so a useful generalist robot must bridge not only a physical generalization gap but a communication gap between how people express intent and how the policy expects it to be specified.

We study this human--VLA communication gap through two mismatches. A natural specification leaves some of the executable task structure unstated, which the policy must recover: an \emph{information gap}. And even a complete specification may fail if its resolved events are organized or invoked outside the policy's learned execution interface: an \emph{execution gap}. Quantifying them gives a human-centered measure of generalist capability, how reliably a non-expert expressing the task naturally can reach it, and a diagnostic account that a success rate alone cannot give: \emph{which} executable structure each form of human specification leaves unresolved, and what each costs the policy.

Studying this gap is difficult for three reasons. First, natural human task specification is open-ended: the same objective can be communicated through direct commands, desired outcomes, or partial references, and no fixed list of paraphrases captures it. Rather than enumerating that space, we begin from how people actually instruct another human. We collect natural instructions from participants shown task demonstrations and, informed by these and by prior work on situated instruction, develop a taxonomy that factors any specification by what task content it states, how it refers to objects, and how it composes multiple manipulations. The taxonomy does two things. It lets us generate human-like
instructions for any demonstrated task, varying what is made explicit
while holding the physical objective fixed. And it lets us measure what
people naturally state, compress, or leave implicit as tasks grow more
complex.

Second, resolving a human specification requires a policy-facing reference for what a fully explicit command looks like. Although VLAs inherit large vision--language backbones, their robot-training supervision is dominated by concise, action-centric instructions with regular structure in actions, object references, and destination descriptions \cite{wanna2026limited}. We therefore mine recurring forms from robot-training datasets and use them to render each grounded event as an explicit action--object--destination command. These policy-facing realizations provide a common reference against which human specifications can be compared, and allow compound objectives to be expressed as sequences of atomic events when we study execution after the information gap has been removed.

Third, a success rate only says that a policy failed, but does not locate the gap. We introduce an \emph{information-gap analysis} that characterizes, for each specification, which executable variables are explicitly provided, which require further grounding, and whether compound manipulations are decomposed into individual events. We compare each specification against a fully grounded command that explicitly identifies every action, target object, and target placement, and execute the same physical tasks under each form, measuring task success and partial progress. This allows us to relate the type and amount of unresolved structure to the VLA's ability to recover the intended behavior. Once this information is fully resolved, we then probe the residual \emph{execution gap} by holding the grounded event sequence fixed while varying whether events are issued together or sequentially, and whether each event begins from the preceding or home robot state. This separates task recovery from execution once the task is fully specified.

Together, our study separates \emph{what a VLA can physically do} from \emph{what a human can reliably ask it to do}, and finds them mismatched in a specific way. As tasks grow, people increasingly rely on contextual grounding and compress multiple manipulations into higher-level expressions, leaving more executable structure implicit, whereas the evaluated policies tolerate grounding dependence comparatively well but pay much more for genuinely missing task semantics and implicit execution structure, especially on complex tasks. We observe the same qualitative trends on $\pi_{0.5}$ \cite{black2025pi05} and $\pi_0$-FAST \cite{pertsch2025fast}, with similar behavior in Cosmos3-Edge \cite{aditi2026cosmos3}, a world-action model. Nor does resolving the information gap guarantee reliable execution: even fully specified atomic events are more reliable when delivered one at a time and invoked from a familiar state.

\section{Related Work}

\subsection{Human Objective Specification and Situated Instruction}
Natural task-oriented communication extends beyond direct imperative commands: speakers express intended acts indirectly \cite{searle1975indirect,clark1980polite}, communicate purpose or constraints \cite{di-eugenio-1992-understanding,vander-linden-di-eugenio-1996-corpus}, and rely on shared perceptual context to identify referents \cite{clark1986referring}, and human--robot dialogue corpora similarly combine action semantics, spatial reference, and interaction context \cite{bastianelli2014huric,zarriesz2016pentoref,lukin2024scout}. We treat these as alternative ways of specifying a physical objective: our taxonomy characterizes what executable information is stated, how referents are grounded, and how manipulations are composed, so that we can measure both how people compress objectives and how policies respond to the resulting implicit structure.

\subsection{The Learned Language Interface of VLA Policies}
Robot-training language is substantially narrower than unrestricted human communication. Datasets such as RT-1 \cite{brohan2023rt1}, BridgeData V2 \cite{walke2023bridgedata}, TacoPlay \cite{rosete2023tacorl}, Language Table \cite{lynch2023interactive}, and LIBERO \cite{liu2023libero} exhibit highly repetitive, dataset-specific instruction distributions \cite{wanna2026limited}, unlike richer embodied-language datasets such as ALFRED \cite{shridhar2020alfred} and SCOUT \cite{lukin2024scout}. Recent studies show that VLA behavior can be sensitive to meaning-preserving linguistic variation \cite{kim2026liberopara,hou2026langgap,fang2025intention,yin2026semanticrebinding}, but largely study variation in already explicit instructions. We focus on a preceding problem, natural specifications that leave task semantics or execution structure implicit, and use the regularities of the policy's fine-tuning language only to construct consistent policy-facing realizations once this information is resolved.

\subsection{Human-Centered Evaluation of Generalist Robot Policies}
Recent benchmarks increasingly move beyond aggregate task success to isolate visual, relational, procedural, and linguistic competencies. RoboLab probes controlled manipulation capabilities and instruction specificity \cite{robolab2026}, while LIBERO-Para and related work study linguistic and semantic robustness under controlled task variations \cite{kim2026liberopara,fang2025intention,hou2026langgap}. Our evaluation asks a complementary question: how reliably can a learned capability be reached from the ways a non-expert naturally specifies the task? For a fixed objective, we measure the \emph{information gap} between a specification and its executable structure, then the residual \emph{execution gap} once that structure is resolved.

\section{Method}

We study the communication gap around a common grounded representation of the physical task: we recover its sequence of manipulation events, characterize how humans naturally specify these events, and construct fully resolved policy-facing realizations of the same task. This shared representation lets us measure what a human specification leaves unresolved, and then test whether the fully resolved task remains sensitive to command organization and invocation state.

\subsection{Problem Formulation}

We consider a physical task $\tau$ demonstrated by a video
$V_\tau=(I_1,\ldots,I_T),$ where $I_t$ is the RGB observation at time $t$. From the demonstration, we recover an ordered sequence of grounded manipulation events $E_\tau=(e_1,\ldots,e_{K_\tau})$, $e_k=(A_k,O_k,D_k)$, where $A_k$ denotes the physical action, $O_k$ the manipulated object, and $D_k$ its destination or resulting target state. For example, placing a banana and an apple into a bowl yields two events. $E_\tau$ is the common semantic reference throughout: human specifications may selectively state, abstract, or compose information from these events, whereas robot-oriented commands express the same events using forms observed in robot-training data, so all variants preserve the same physical objective and differ only in how it is communicated. For each task, let $\mathcal{H}_\tau=\{u_\tau^1,\ldots,u_\tau^N\}$ denote its human-like specifications, and $R_\tau$ the fully grounded robot-oriented realization of the same event sequence.

Given a fixed VLA policy $\pi$, the \emph{information gap} is characterized by what structure in $E_\tau$ is not directly available from a human specification and must be recovered for execution; once resolved, the residual \emph{execution gap} asks whether the same executable task remains reliable as its events are organized differently and invoked from different robot states. We recover $E_\tau$ from demonstration videos using object-centric tracking and persistent changes in state and relations, with a VLM providing factual descriptions of the extracted transitions for language generation (Appendix~\ref{app:event_extraction}).

\subsection{Characterizing Human Objective Specification}
\label{sec:human}

\textbf{Human Instruction Elicitation.}
Natural objective specification is open-ended, so rather than defining this space a priori, we first study how people instruct another human from demonstrated behavior. Participants observe task demonstrations of varying complexity and write the instruction they would give another human to achieve the demonstrated objective. We prescribe no command format, so participants decide what to state explicitly, how to identify relevant entities, and how to express compound objectives. These instructions provide both evidence for constructing our taxonomy and a natural distribution against which its forms can be analyzed.

 \begin{table*}[t]
\centering
\scriptsize
\setlength{\tabcolsep}{4pt}
\renewcommand{\arraystretch}{1.05}
\begin{tabular}{
    L{0.115\textwidth}
    L{0.095\textwidth}
    L{0.14\textwidth}
    L{0.58\textwidth}
}
\toprule
\textbf{Factor} & \textbf{Form} & \textbf{Controlled variation} & \textbf{Example} \\
\midrule
\multirow{3}{*}{\textbf{Task content}}
& Action & Action performed &
\ex{``\vary{Put} all the yellow objects in the left bin and all the blue objects in the right bin.''} \\
& Outcome & Desired physical goal state &
\ex{``All yellow objects \vary{should be in} the left bin, and all blue objects \vary{should be in} the right bin.''} \\
& Intent / preference & Desired effect or preference &
\ex{``\vary{I want the objects sorted by color}.''} \\
\midrule
\multirow{3}{*}{\textbf{Reference form}}
& Direct & Directly named referent &
\ex{``Put the yellow mug in \vary{the left bin} and the blue cup in \vary{the right bin}.''} \\
& Grounding-dependent & Scene-resolved referent &
\ex{``Put the yellow mug in \vary{the bin on the left} and the blue cup in \vary{the bin on the right}.''} \\
& Missing & Referent omitted &
\ex{``Put the yellow mug and the blue cup away.'' \textit{(destination omitted)}} \\
\midrule
\multirow{3}{*}{\textbf{Composition}}
& Enumerated & One manipulation per clause &
\ex{``Put \vary{the yellow mug} in the left bin, then put \vary{the yellow bowl} in the left bin, then put \vary{the blue cup} in the right bin.''} \\
& Grouped & Shared predicate across objects &
\ex{``Put \vary{the yellow mug and bowl} in the left bin, and \vary{the blue cup and plate} in the right bin.''} \\
& Collective & Set-level description &
\ex{``Put \vary{all the yellow objects} in the left bin and \vary{all the blue objects} in the right bin.''} \\
\bottomrule
\end{tabular}
\caption{Factorized taxonomy of human objective specification. All example variants describe the same physical task; highlighted spans mark the varied element.}
\label{tab:human_taxonomy}
\end{table*}
 
\textbf{Deriving the Specification Taxonomy.}
We analyze the collected instructions together with distinctions from prior work on task-oriented and situated language \cite{searle1975indirect,clark1986referring,di-eugenio-1992-understanding,vander-linden-di-eugenio-1996-corpus}. Three recurring factors capture the major ways specifications differ (Table~\ref{tab:human_taxonomy}): \emph{task content}, whether the speaker states an action, a desired physical outcome, or an intent or preference from which the physical objective must be inferred; \emph{reference form}, whether relevant entities are named directly, identified through scene-dependent descriptions, or omitted altogether; and \emph{composition}, whether multiple manipulations are enumerated individually, grouped under a shared predicate, or compressed into a collective description. The taxonomy serves two roles: descriptively, to code the collected human instructions and measure how the information people provide changes with task complexity; and experimentally, to generate controlled specification variants for VLA evaluation.

\textbf{Taxonomy-Conditioned Specification Generation.}
Given the grounded event sequence $E_\tau$, we generate human-like specifications by selecting a configuration of the three factors: a \emph{content operator} (action, resulting physical state, or higher-level intent), a \emph{reference operator} (direct naming, or description through attributes, spatial relations, or semantic groupings), and a \emph{composition operator} (separate, grouped under a shared predicate, or summarized collectively). A language model then produces lexical and syntactic variants without changing the selected configuration, providing controlled specifications with known executable structure while keeping the physical objective fixed.

\subsection{Characterizing the VLA Language Interface}
\label{sec:VLA-side}

\textbf{Robot-Training Language Prior.}
To characterize the other side of the interface, we model how fully resolved manipulation events are expressed in robot-training data: unlike the human space, which varies what task structure is made explicit, the policy-facing space keeps the grounded task fixed and varies only the linguistic forms through which the VLA was exposed to actions, objects, and destinations during training. For each evaluated policy, we mine the robot-language dataset corresponding to its fine-tuning distribution. Since robot-training language is substantially narrower and more repetitive than unrestricted human language \cite{wanna2026limited}, we extract from each instruction its command template with the action, manipulated-object reference, and destination reference, replacing grounded entities with typed placeholders to normalize commands with the same structure. These recurring forms are a language prior for constructing policy-facing commands compatible with the policy's fine-tuning distribution (Appendix~\ref{app:robot_language}).

\textbf{Policy-Facing Realization.}
Given a grounded event $e_k=(A_k,O_k,D_k)$, we instantiate a compatible mined template with its action, object, and destination to obtain a fully explicit policy-facing command $r_k$; for a compound task, this yields an atomic command sequence $R_\tau=(r_1,\ldots,r_{K_\tau}).$ These realizations are a fully resolved instruction compatible with the policy's training interface: the reference against which human specifications are compared and, for compound tasks, the atomic commands used to study execution. We keep them fixed so that performance differences reflect unresolved task information or execution conditions rather than incidental linguistic variation.

\subsection{Measuring the Communication Gap}
\label{sec:measuring_gap}

We measure the \emph{information gap} by comparing each specification with the grounded event sequence $E_\tau$, capturing what task information is available and how explicitly its events are decomposed. We then measure the residual \emph{execution gap} by varying command organization and invocation state after the task is fully resolved.

\textbf{Semantic Coverage and Grounding Dependence.}
For each event $e_k=(A_k,O_k,D_k)$, let $\mathcal{S}_k=\{A_k,O_k,D_k\}$. For every required component $s\in\mathcal{S}_k$, we assign $\sigma_{k,s}(u)\in\{0,1,2\},$
where $\sigma_{k,s}(u)=0$ if the component is omitted from specification $u$, $\sigma_{k,s}(u)=1$ if it is specified but its grounded value must be resolved from the scene/task context, and $\sigma_{k,s}(u)=2$ if its grounded value is directly specified.

We define the \emph{semantic coverage ratio}
\begin{equation}
\eta_\tau(u) := \frac{
\sum_{k=1}^{K_\tau}
\sum_{s\in\mathcal{S}_k}
\mathbbm{1}[\sigma_{k,s}(u)\geq1]
}{
3K_\tau
},\label{equ:coverage}
\end{equation}
which measures the fraction of required task components represented in the specification, either directly or through a grounding-dependent expression. We separately define the \emph{groundedness ratio}
\begin{equation}
\gamma_\tau(u) := \frac{
\sum_{k=1}^{K_\tau}
\sum_{s\in\mathcal{S}_k}
\mathbbm{1}[\sigma_{k,s}(u)=2]
}{
3K_\tau
},\label{equ:groundedness}
\end{equation}
which measures the fraction whose grounded values are directly given in language. Thus, $1-\eta_\tau(u)$ captures omitted task information, while $\eta_\tau(u)-\gamma_\tau(u)$ captures information that is present but still requires grounding.

\textbf{Task Decomposition.}
Semantic coverage does not capture whether a compound objective is explicitly organized into its constituent manipulations. Let $J_\tau(u)$ denote the number of distinct manipulation units individuated in specification $u$. We define the decomposition ratio
$\delta_\tau(u):=\frac{J_\tau(u)}{K_\tau}.$ A value of $\delta_\tau(u)=1$ indicates that every atomic event is separately expressed, while smaller values indicate that multiple events are compressed into a shared expression.

\textbf{Reasoning Required to Resolve the Specification.}
The same amount of unresolved structure may require qualitatively different forms of recovery. We additionally categorize specifications by the reasoning needed to map them onto $E_\tau$, including visual grounding, action inference, task decomposition, and intent inference. We use this as a complementary analysis of \emph{what computation} is required to close the information gap; details are provided in Appendix~\ref{app:reasoning_analysis}.

\textbf{Information-Gap Evaluation.}
For each specification $u$, we execute the corresponding physical task with a fixed VLA policy and measure task success and partial progress. We compare it against a matched fully resolved realization $r_\tau^{\mathrm{full}}$ constructed from the same event sequence $E_\tau$. This reference uses Action--Direct--Enumerated semantics: every required action, object, and destination is directly specified, and every manipulation event is explicitly individuated. All events are presented together in a single prompt, so this comparison resolves the task information. For evaluation metric $M$, we define the behavioral information gap as
\begin{equation}
\Delta_M^{\mathrm{info}}(u)
=
M(r_\tau^{\mathrm{full}})-M(u).
\end{equation}
A larger $\Delta_M^{\mathrm{info}}(u)$ indicates a larger performance cost from executable structure left unresolved in $u$. We analyze this cost primarily as a function of semantic coverage
$\eta_\tau(u)$, groundedness $\gamma_\tau(u)$, decomposition
$\delta_\tau(u)$, and task complexity. A complementary analysis by
required reasoning operation is provided in Appendix~\ref{app:reasoning_analysis}. Together, these analyses characterize both \emph{how much} executable structure is unresolved and \emph{which kind} the policy struggles to recover.

\textbf{Residual Execution Gap.}
We next ask whether resolving the information gap is sufficient for reliable execution. All conditions use the same fully specified Action--Direct--Enumerated event sequence and the same policy-facing atomic commands $R_\tau=(r_1,\ldots,r_{K_\tau})$. We vary only how these commands are executed. First, we compare presenting all atomic events together in one prompt with issuing the same events sequentially, one command at a time. This isolates sensitivity to \emph{execution organization} while holding the task semantics fixed. For metric $M$, we measure
\begin{equation}
\Delta_M^{\mathrm{org}}
=
M(R_\tau^\text{ sequential})
-
M(R_\tau^\text{ one prompt}).
\end{equation}

Second, under sequential execution, we vary the robot state from which each subsequent command is invoked. In the chained condition, each command begins from the terminal robot configuration left by the preceding event. In the reset condition, the arm is returned to its home configuration before the next command while preserving the task state produced by previous events. We measure
\begin{equation}
\Delta_M^{\mathrm{state}}
=
M(R_\tau^\text{ sequential + home reset})
-
M(R_\tau^\text{ sequential}).
\end{equation}
The home reset is not proposed as a solution, but serves as a generic intervention for testing sensitivity to invocation state. Together, these comparisons separate information recovery from execution failure. Full execution and evaluation details are provided in Appendix~\ref{app:evaluation_protocol}.
\section{Experiments}
\label{sec:experiments}

We investigate three questions about the human--VLA communication gap. \textbf{Q1}: How do natural human specifications change with task complexity, and what task information do people compress or leave implicit? \textbf{Q2}: Which forms of unresolved structure can the VLA recover, and how does their cost change as tasks become more complex? \textbf{Q3}: Once the task information is fully resolved, what residual execution gap remains in how the resolved events are presented and invoked?

\subsection{Experimental Setup}

\textbf{Environment and Policies.}
We evaluate $\pi_{0.5}$ \cite{black2025pi05}, $\pi_0$-FAST \cite{pertsch2025fast}, and Cosmos3-Edge \cite{aditi2026cosmos3}, using checkpoints fine-tuned on DROID, on 120 RoboLab tasks with 10 rollouts for each task--instruction condition. All three policies exhibit similar qualitative trends across our information- and execution-gap analyses. For clarity, the main text focuses on $\pi_{0.5}$, which achieves the strongest overall task performance and therefore provides the clearest dynamic range for measuring the communication gap. Results for $\pi_0$-FAST and Cosmos3-Edge are reported in Appendix~\ref{app:additional_models}. 

\textbf{Human Instruction Study.}
To study how people naturally communicate physical objectives, we select 15 tasks spanning single-event manipulations and compound objectives with different structural regularities. Fifty participants view each demonstration without its canonical instruction and write the instruction they would naturally give to another human, yielding 750 specifications. We use these annotations to analyze how specification length, semantic coverage, grounding dependence, and decomposition change with task complexity.

\textbf{Task Complexity and Compression Affordance.}
We measure physical task complexity by the number of grounded atomic events $K_\tau$. Because tasks with the same event count may differ substantially in how naturally they can be compressed, we additionally distinguish compound tasks that require event-wise enumeration, admit a compact grounded rule, or admit a higher-level task effect. These labels are assigned from the grounded objective independently of the human-written specifications. Full definitions and the RoboLab-120 distribution are provided in Appendix~\ref{app:task_complexity}

\begin{table*}[t]
\centering
\scriptsize
\setlength{\tabcolsep}{3pt}
\renewcommand{\arraystretch}{1.15}
\begin{tabular}{
L{0.075\linewidth}
L{0.11\linewidth}
C{0.035\linewidth}
C{0.035\linewidth}
C{0.095\linewidth}
C{0.10\linewidth}
C{0.095\linewidth}
C{0.10\linewidth}
C{0.095\linewidth}
C{0.10\linewidth}
}
\toprule
\multirow{3}{*}{\textbf{Content} $\downarrow$} &
\multirow{3}{*}{\textbf{Reference} $\downarrow$} &
\multirow{3}{*}{$\boldsymbol{\eta_\tau}$} &
\multirow{3}{*}{$\boldsymbol{\gamma_\tau}$} &
\multicolumn{6}{c}{\textbf{Composition} $\rightarrow$} \\
\cmidrule(lr){5-10}
 & & & &
\multicolumn{2}{c}{\cellcolor{structblue!35}\begin{tabular}[c]{@{}c@{}}Enumerated\\($\bar{\delta}{=}1.00$, $N{=}120$)\end{tabular}} &
\multicolumn{2}{c}{\cellcolor{structblue!16}\begin{tabular}[c]{@{}c@{}}Grouped\\($\bar{\delta}{=}0.47{\pm}0.08$, $N{=}54$)\end{tabular}} &
\multicolumn{2}{c}{\cellcolor{structblue!14}\begin{tabular}[c]{@{}c@{}}Collective\\($\bar{\delta}{=}0.41{\pm}0.04$, $N{=}50$)\end{tabular}} \\
 & & & &
Success (\%) & $\Delta_{\mathrm{Success}}^{\mathrm{info}}$ &
Success (\%) & $\Delta_{\mathrm{Success}}^{\mathrm{info}}$ &
Success (\%) & $\Delta_{\mathrm{Success}}^{\mathrm{info}}$ \\
\midrule
Action & Direct & \cellcolor{structblue!35}1.00 & \cellcolor{structblue!35}1.00 &
38.0 $\pm$ 5.4 & 0.0 (ref.) &
15.9 $\pm$ 6.0 & \cellcolor{gapred!23}\textbf{7.6 $\pm$ 3.1} &
13.1 $\pm$ 5.8 & \cellcolor{gapred!31}\textbf{10.4 $\pm$ 3.5} \\
 & Grounding-dep. & \cellcolor{structblue!35}1.00 & \cellcolor{structblue!12}0.33 &
35.2 $\pm$ 5.3 & \cellcolor{gapred!8}\textbf{2.8 $\pm$ 1.8} &
13.5 $\pm$ 5.6 & \cellcolor{gapred!30}10.0 $\pm$ 3.3 &
12.4 $\pm$ 5.7 & \cellcolor{gapred!33}11.1 $\pm$ 3.6 \\
 & Missing & \cellcolor{structblue!12}0.33 & \cellcolor{structblue!12}0.33 &
26.9 $\pm$ 4.9 & \cellcolor{gapred!33}\textbf{11.1 $\pm$ 2.3} &
7.1 $\pm$ 4.2 & \cellcolor{gapred!49}16.4 $\pm$ 3.9 &
6.5 $\pm$ 4.2 & \cellcolor{gapred!51}17.0 $\pm$ 4.1 \\
\midrule
Outcome & Direct & \cellcolor{structblue!23}0.67 & \cellcolor{structblue!23}0.67 &
33.5 $\pm$ 5.2 & \cellcolor{gapred!14}\textbf{4.5 $\pm$ 1.9} &
11.8 $\pm$ 5.3 & \cellcolor{gapred!35}11.7 $\pm$ 3.5 &
9.7 $\pm$ 5.1 & \cellcolor{gapred!41}13.8 $\pm$ 3.8 \\
 & Grounding-dep. & \cellcolor{structblue!23}0.67 & 0.00 &
31.6 $\pm$ 5.2 & \cellcolor{gapred!19}6.4 $\pm$ 2.0 &
10.0 $\pm$ 5.0 & \cellcolor{gapred!41}13.5 $\pm$ 3.7 &
9.2 $\pm$ 5.0 & \cellcolor{gapred!43}14.3 $\pm$ 3.9 \\
 & Missing & 0.00 & 0.00 &
24.6 $\pm$ 4.8 & \cellcolor{gapred!40}13.4 $\pm$ 2.4 &
5.3 $\pm$ 3.7 & \cellcolor{gapred!55}18.2 $\pm$ 4.1 &
4.9 $\pm$ 3.7 & \cellcolor{gapred!56}18.6 $\pm$ 4.3 \\
\midrule
Intent & Direct & \cellcolor{structblue!12}0.33 & \cellcolor{structblue!12}0.33 &
22.8 $\pm$ 4.7 & \cellcolor{gapred!46}\textbf{15.2 $\pm$ 2.6} &
5.3 $\pm$ 3.7 & \cellcolor{gapred!55}18.2 $\pm$ 4.1 &
4.3 $\pm$ 3.5 & \cellcolor{gapred!58}19.2 $\pm$ 4.4 \\
 & Grounding-dep. & \cellcolor{structblue!12}0.33 & 0.00 &
22.1 $\pm$ 4.6 & \cellcolor{gapred!48}15.9 $\pm$ 2.6 &
4.7 $\pm$ 3.5 & \cellcolor{gapred!56}18.8 $\pm$ 4.2 &
4.3 $\pm$ 3.5 & \cellcolor{gapred!58}19.2 $\pm$ 4.4 \\
 & Missing & 0.00 & 0.00 &
18.8 $\pm$ 4.3 & \cellcolor{gapred!58}19.3 $\pm$ 2.8 &
3.2 $\pm$ 2.9 & \cellcolor{gapred!61}20.3 $\pm$ 4.3 &
2.9 $\pm$ 2.9 & \cellcolor{gapred!62}20.6 $\pm$ 4.5 \\
\bottomrule
\end{tabular}
\caption{$\pi_{0.5}$ performance across the full factorial space of taxonomy-controlled specifications.
Less explicit specifications (lighter blue, less decomposed columns) yield lower success and larger information-gap penalties (darker red); bold marks one-factor ablations from the fully specified Action--Direct--Enumerated reference.
Rows vary content and reference; columns vary composition (enumerated, grouped, and collective forms on $N=120$, $54$, and $50$ tasks).
Mean $\pm$ 95\% CI over task-level means after averaging rollouts and surface realizations.
Results for $\pi_0$-FAST and Cosmos3-Edge, which show similar trends, are in Appendix~\ref{app:additional_models}.}
\label{tab:information_gap_main}
\vspace{-10pt}
\end{table*}

\textbf{Controlled VLA Evaluation.}
For all 120 tasks, we generate human-like specifications under controlled taxonomy configurations and evaluate them against the fully grounded $r_\tau^{\text{full}}$. This produces the factorial evaluation in Table~\ref{tab:information_gap_main}, where task content, reference form, and composition systematically vary the executable structure exposed to the policy.
To study the residual execution gap, we then hold the grounded event sequence fixed and vary how its atomic events are organized and when they are issued to the policy. Full execution protocols are provided in Appendix~\ref{app:evaluation_protocol}.


\subsection{Humans Compress as Task Complexity Increases}
\label{sec:human_compression}

We first ask whether natural human communication actually creates the information gap studied in Sec.~\ref{sec:measuring_gap}. If people expressed a compound objective by simply concatenating atomic commands, instruction length and explicit event structure should grow with the number of required manipulations. Instead, we find people increasingly compress the objective, leaving more executable structure implicit.

\begin{figure}[t]
    \centering
\includegraphics[width=\columnwidth]{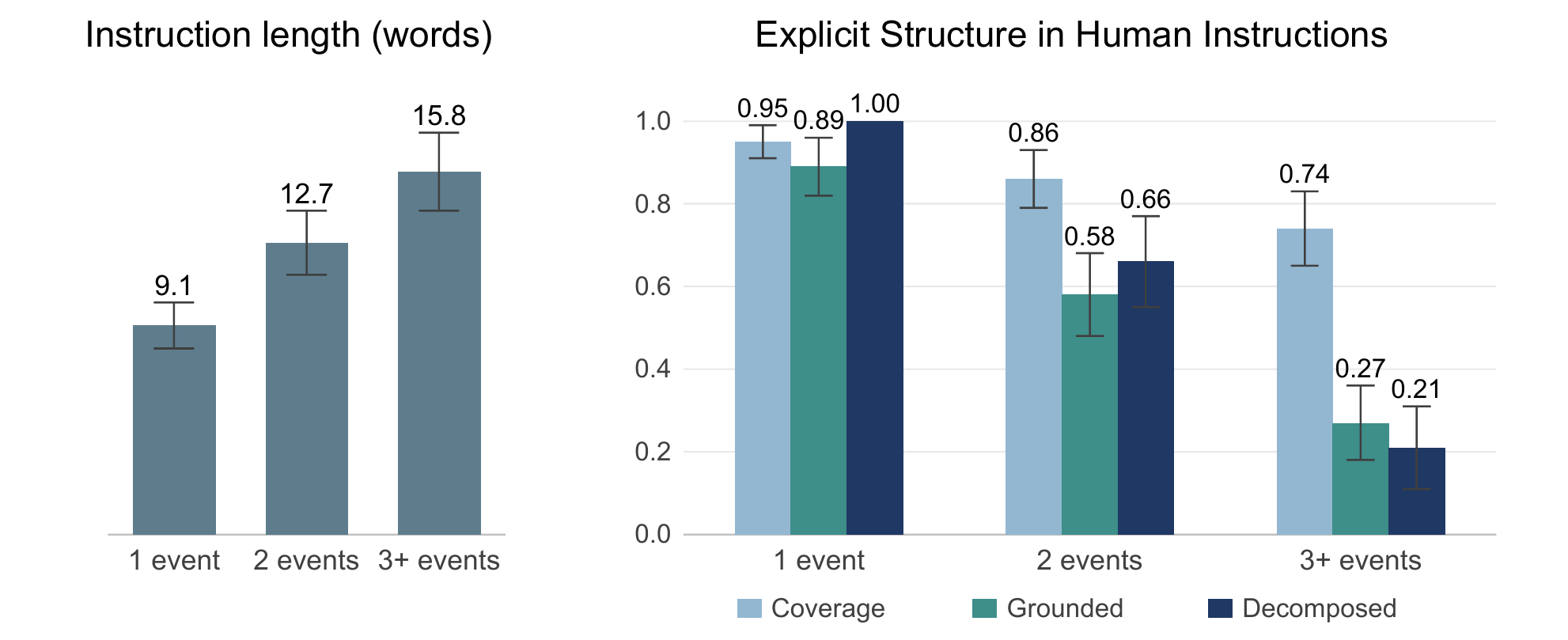}
    \caption{\textbf{Humans increasingly compress execution structure as tasks grow.}
    Instruction length increases with the number of events (left), but the executable structure made explicit decreases (right): coverage declines moderately, while groundedness and decomposition fall sharply.}
\label{fig:human_compression}
\end{figure}

\textbf{Instructions Remain Compact as Task Complexity Grows.}
Fig.~\ref{fig:human_compression} compares natural specifications as the number of required atomic events $K_\tau$ increases. Instruction length grows sublinearly in $K_\tau$, from $9.1$ words for single-event tasks to $15.8$ words for tasks requiring three or more events. At the same time, the executable structure made explicit declines substantially. Semantic coverage $\eta_\tau$ decreases from $0.95$ to $0.74$, while groundedness $\gamma_\tau$ falls from $0.89$ to $0.27$ and decomposition $\delta_\tau$ from $1.00$ to $0.21$. Thus, people do not communicate increasingly complex objectives by proportionally increasing instruction length. They preserve much of the overall task content while increasingly relying on the listener to resolve referents and, especially, to expand a compact expression into atomic manipulations. Natural communication therefore produces a growing information gap as the physical objective becomes more compositional.

\textbf{Compression Exploits Available Task Structure.}
This compression is not simply a consequence of having more events. Tasks with the same $K_\tau$ differ in whether their events admit a compact shared description, and people exploit this structure when it is available. Weakly structured compound objectives remain comparatively action-oriented and enumerative, whereas rule-compressible objectives shift toward outcome-level and collective descriptions. When the task admits a compact semantic effect, participants more often describe the intent itself and leave the concrete physical realization implicit. The full distribution across task-content and composition forms is reported in Appendix~\ref{app:human_strategy}. Together, these results show that human compression is \emph{structure-dependent}: people preserve event-wise detail when the objective requires it, but use increasingly abstract specifications when a compact rule or intent can characterize the same physical goal.

\subsection{VLAs: Selectively Sensitive to Unresolved Task Structure}
\label{sec:information_gap_results}

\begin{figure*}[t]
    \centering
    \includegraphics[width=1\textwidth]{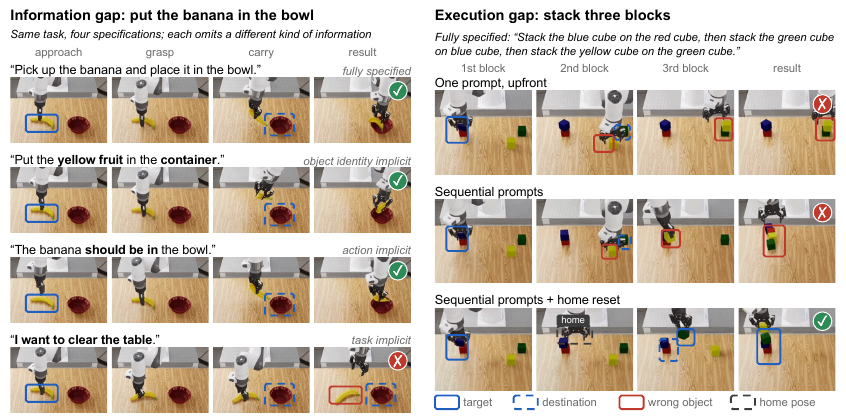}
    \caption{\textbf{Information and execution gaps in human--VLA communication.}
    \textbf{Left:} The policy succeeds when object identity or action is implicit, but fails when task semantics must be inferred.
    \textbf{Right:} Multi-step execution can fail through incorrect object selection; returning to a home pose between subtasks enables completion in the illustrated rollout. Solid blue: targets; dashed blue: destinations; red: incorrect objects; dashed gray: home pose.}
    \label{fig:communication_gap_rollouts}
    \vspace{-10pt}
\end{figure*}

Can the policy recover different unresolved structure equally well? As in the left panel of Fig.~\ref{fig:communication_gap_rollouts}, the physical task is held fixed while the specification leaves different parts implicit; Table~\ref{tab:information_gap_main} evaluates this over the full factorial space of taxonomy-controlled specifications for $\pi_{0.5}$ ($\pi_0$-FAST and Cosmos3-Edge show the same trends, Appendix~\ref{app:additional_models}).

\begin{figure}[t]
    \centering
    \includegraphics[width=\columnwidth]{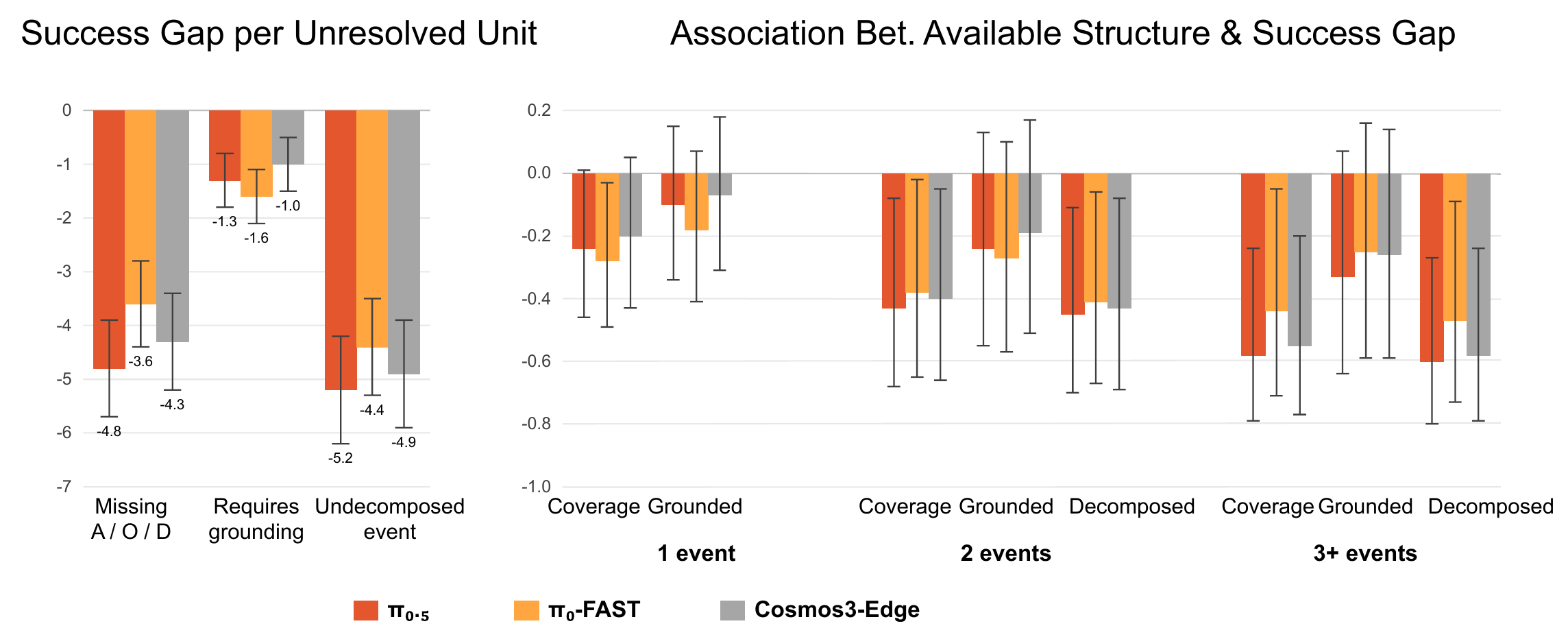}
    \caption{\textbf{Behavioral cost of unresolved task structure.}
    \textbf{Left:} For all three policies, grounding dependence is cheap, while missing action/object/destination information and undecomposed events cost far more.
    \textbf{Right:} With task complexity, coverage and decomposition become more strongly associated with the success gap than groundedness.}
    \label{fig:information_gap_behavior}

\end{figure}

\begin{figure}[t]
    \centering
    \includegraphics[width=\columnwidth]{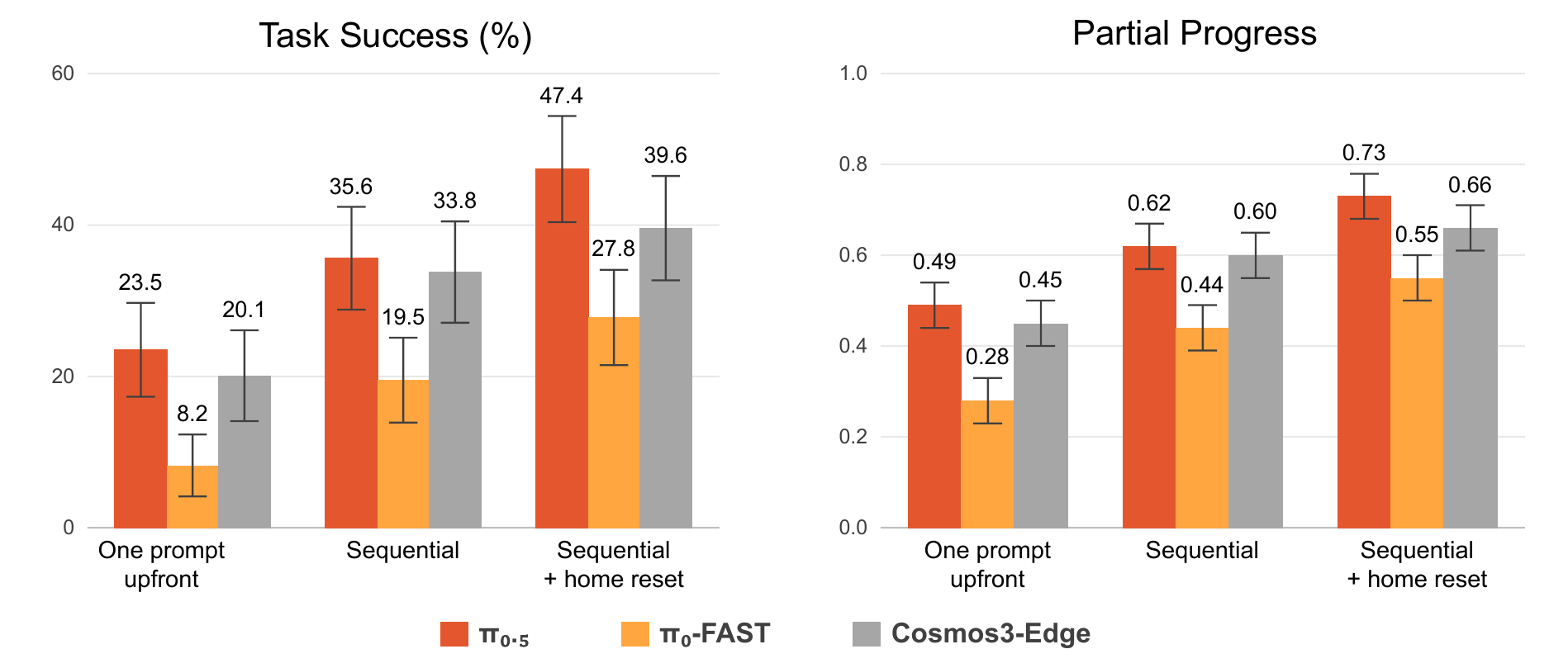}
    \caption{\textbf{Residual execution gap.}
    On the same fully specified Action--Direct--Enumerated tasks, issuing atomic events sequentially improves success and partial progress over a single upfront prompt for all three policies, and resetting the arm to its home configuration between events adds a further gain.}
    \label{fig:observation_alignment_gap}

\end{figure}

\textbf{What matters is not simply how much information is unresolved, but what must be recovered.}
The factorial design isolates different forms of unresolved structure while
holding the remaining specification maximally explicit.
Along the task-content axis, replacing an action specification with an outcome
specification produces only a $4.5$-point matched success gap, whereas an
intent-level specification increases the gap to $15.2$ points.
An explicit outcome therefore often constrains the omitted action sufficiently
for the policy to recover it, while an intent leaves the physical goal itself
to be inferred before any executable behavior can be determined.
The same asymmetry appears in reference: with action content and enumerated
composition fixed, grounding-dependent references incur only a $2.8$-point
gap, whereas missing reference information incurs an $11.1$-point gap.
Thus, the policies are substantially more robust to \emph{resolution} of
information already constrained by language and observation than to
\emph{reconstruction} of task structure that is not present in the
specification.

\textbf{Missing event structure is as costly as missing task semantics.}
With action content and direct references fixed, grouping manipulations creates a
$7.6$-point success gap and collective specification a $10.4$-point gap.
Across all three policies, each undecomposed event costs $4.4$--$5.2$ success
points, comparable to a missing action, object, or destination
($3.6$--$4.8$), and far more than a component that only requires grounding
($1.0$--$1.6$).
Thus, even when task semantics are available, VLAs remain sensitive to whether
the objective is explicitly decomposed into executable events.

\textbf{Structural reconstruction becomes harder as tasks grow.}
As task complexity increases, semantic coverage $\eta_\tau$ and decomposition
$\delta_\tau$ become more strongly associated with the success gap, while groundedness $\gamma_\tau$ remains weaker
(Fig.~\ref{fig:information_gap_behavior}, right).
This creates a growing mismatch with human communication: as task complexity grows, human decomposition drops
(Fig.~\ref{fig:human_compression}), precisely as policies become more sensitive
to missing event structure.
The communication gap widens from both sides: people expose less
structure as tasks grow, while VLAs become less able to recover it.

\subsection{Execution Gap After Information Resolution}
\label{sec:alignment_gap_results}

We next characterize what executable information the
policy needs. We next ask whether providing that information completely
is sufficient for reliable execution. As illustrated in the right panel
of Fig.~\ref{fig:communication_gap_rollouts}, we fix every compound task
to the same fully resolved Action--Direct--Enumerated event sequence and
vary only how those events are presented and invoked. This isolates a
residual execution gap after task semantics, grounding, and event
structure have already been resolved.

\textbf{Sequentializing resolved events substantially improves execution.}
Is complete executable information sufficient for reliable execution? As in the right panel of Fig.~\ref{fig:communication_gap_rollouts}, we fix every compound task to the same fully resolved Action--Direct--Enumerated event sequence and vary only how its events are presented and invoked. Across all three policies, issuing the grounded events sequentially outperforms presenting them in one upfront prompt in both success and partial progress (Fig.~\ref{fig:observation_alignment_gap}). No task information is added; the difference is execution organization: a compound prompt requires several manipulations from one invocation, whereas sequentialization reduces each invocation to one atomic event. Resolving the task semantics is thus not enough; the objective must also be exposed at a granularity the policy can reliably execute.

\textbf{Invocation state creates a second residual gap.}
In sequential execution, later commands are invoked from robot configurations produced by preceding actions, often far from the states seen during training; resetting the arm to its home configuration before each event makes the invocation observation more in-distribution and improves success for all three policies (Fig.~\ref{fig:observation_alignment_gap}). Even with the correct atomic command, execution can thus fail because the policy is invoked from an unfamiliar state: each behavior must be accessed from an observation distribution in which it remains reliable.

\subsection{Three Policies, One Gap, Different Weak Points}

Across all
three policies, we observe that grounding dependence is cheap,
missing semantics and undecomposed events are expensive, and sequential
invocation from a familiar state improves execution. However, within that shared
pattern, the models differ in ways that track their architecture.

\textbf{The information gap is shaped by the language backbone.}
Cosmos3-Edge pays the least for grounding-dependent references, consistent with a general-purpose
vision--language backbone that resolves attribute and spatial reference. $\pi_0$-FAST pays the most ($-1.6$), and its penalties compress overall
because it operates near floor on compound tasks. The decomposition penalty, however,
is the largest term for every model, including $\pi_{0.5}$, whose explicit
subtask prediction narrows but does not close it. No evaluated architecture
recovers implicit event structure reliably.

\textbf{The execution gap splits by mechanism.} The sequentialization gain
is shared, and largest for $\pi_0$-FAST (8.2\% to 19.5\%): it reflects the
short, single-event commands common to all DROID-trained policies, not any
particular action head. The invocation-state gain is not shared. Cosmos3-Edge
gains 5.8 points from a home reset against 11.8 for $\pi_{0.5}$. A
world-action model denoises its action chunk jointly with an imagined
rollout from the current frame, so an unfamiliar arm configuration is
partly absorbed by the video prior rather than falling outside the action
distribution.

\textbf{Implication.} The communication gap is not one property of a single
policy. Its grounding component is inherited from the vision--language
backbone, its organization component from the robot-language distribution,
and its invocation-state component from the action-generation paradigm.
Only the structural-reconstruction component, the one humans exercise most
as tasks grow, is unaddressed by any of
the three, which marks it as the primary target for policy learning.
\section{Conclusion}

We identify a human--VLA communication gap between how people naturally specify tasks and how current policies can reliably act on them, localizing where communication breaks down and what structure must be recovered as concrete targets for policy learning, evaluation, and interface design. Ultimately, generalist robots should adapt to how people communicate, rather than requiring people to adapt to the robot's learned interface.


\bibliographystyle{IEEEtran}
\bibliography{references}

\onecolumn
\newpage
\appendix

\subsection{Task-Structure Complexity and Compression Affordance}
\label{app:task_complexity}

Our analysis distinguishes the \emph{number of required physical events} from the \emph{structure available for expressing those events compactly}. Two tasks with the same number of manipulations may afford very different specifications: unrelated manipulations may require event-wise enumeration, whereas several placements may be summarized by a rule such as ``put all the black objects in the bin.''

\textbf{Atomic Event Count.}
For each task $\tau$, we use the grounded event sequence
\begin{equation}
E_\tau=(e_1,\ldots,e_{K_\tau}),
\end{equation}
where each $e_k=(A_k,O_k,D_k)$ denotes one object-centric physical transition. $K_\tau$ therefore measures physical manipulation complexity at the same granularity used throughout our information-gap analysis. It is derived from the extracted demonstration rather than from instruction length or linguistic structure. This differs from RoboLab's internal \texttt{num\_subtasks} statistic, which is defined by the benchmark's success-condition representation \cite{robolab2026}. We use the latter only as auxiliary metadata and compute $K_\tau$ from $E_\tau$.

\textbf{Compression Affordance.}
Event count alone does not determine how compactly an objective can be communicated. For tasks with $K_\tau>1$, we classify the strongest task-level structure that permits the objective to be expressed without enumerating every event. Classification is performed from $E_\tau$ and the corresponding grounded scene, independently of the human-written specifications. \emph{Enumerative} tasks require event-specific object, destination, or ordering information to remain explicit. Even when events share a generic action or destination, no single non-enumerative rule fully determines the required event sequence. \emph{Rule-compressible} tasks admit a grounded rule $P_\tau$ over observable entities, attributes, relations, or target states such that $E_\tau=\mathcal{E}(P_\tau,G_\tau),$ where $\mathcal{E}$ instantiates the atomic events satisfying $P_\tau$ in scene $G_\tau$. Examples include `all black objects in the bin,'' `all mugs upright,'' or sorting objects by color. \emph{Intent-compressible} tasks admit a higher-level semantic effect or purpose $F_\tau$ that compactly characterizes the objective without directly specifying the grounded object--destination assignments. Recovering the executable task therefore requires intent-to-goal inference: $E_\tau=\mathcal{I}(F_\tau,G_\tau),$ where $\mathcal{I}$ maps the task-level effect and current scene to the corresponding grounded events. Examples include clearing a work surface, packing objects for storage, or discarding unwanted items. Table~\ref{tab:task_structure_definition} summarizes the operational criteria.

\textbf{RoboLab-120 Distribution.}
We apply this coding scheme to the 120 RoboLab tasks \cite{robolab2026}. Table~\ref{tab:robolab_task_structure} reports the resulting distribution. Nearly half of RoboLab tasks require multiple object-centric manipulations, but most compound objectives also exhibit reusable structure. RoboLab therefore contains many cases in which physical complexity increases while the objective remains linguistically compressible, allowing us to study whether people exploit this structure when deciding what to state explicitly and what to leave for the listener to recover.

\begin{table*}[h]
\centering
\footnotesize
\setlength{\tabcolsep}{4pt}
\begin{tabular}{
L{0.13\linewidth}
L{0.25\linewidth}
L{0.21\linewidth}
>{\raggedright\arraybackslash\scriptsize}p{0.31\linewidth}
}
\hline
\textbf{Task structure} &
\textbf{Operational criterion} &
\textbf{Information retained by compact form} &
\textbf{RoboLab examples} \\
\hline
Atomic &
$K_\tau=1$ &
One grounded $(A,O,D)$ event &
\texttt{BananaInBowlTask}; \texttt{BowlInBinTask} \\
\hline
Compound, enumerative &
$K_\tau>1$ and critical event-specific arguments or ordering must still be enumerated &
Individual object identities and/or event ordering &
\texttt{RubiksCubeAndBananaTask}; \texttt{BananaThenRubiksCubeTask}; \texttt{RubiksCubeThenBananaTask} \\
\hline
Compound, rule-compressible &
$K_\tau>1$ and a grounded rule $P_\tau$ generates the required events without event-wise enumeration &
Observable set membership, relation, shared state, or attribute--destination mapping &
\texttt{BlackItemsInBinTask}; \texttt{FoodPackingByColorTask}; \texttt{ReorientAllMugsTask}; \texttt{BlockStackingSpecifiedOrderTask} \\
\hline
Compound, intent-compressible &
$K_\tau>1$ and a semantic effect $F_\tau$ compactly characterizes the objective while leaving concrete event structure to be inferred &
Desired purpose or task-level effect rather than complete grounded $(O,D)$ assignments &
\texttt{CleanUpToysTask}; \texttt{CookingClearPlateTask}; \texttt{RecycleCartonTask}; \texttt{ThrowAwaySnacksTask} \\
\hline
\end{tabular}
\caption{Operational definition of task structure used to analyze human objective specification. Event count measures physical manipulation complexity, whereas compression affordance characterizes the structure available for expressing those events compactly.}
\label{tab:task_structure_definition}
\end{table*}
\begin{table}[h]
\centering
\footnotesize
\begin{tabular}{
p{0.55\linewidth}
p{0.14\linewidth}
p{0.18\linewidth}
}
\hline
\textbf{Task structure} &
\textbf{Tasks} &
\textbf{Fraction (\%)} \\
\hline
Atomic &
64 &
$53.3\%$ \\
Compound, enumerative &
3 &
$2.5\%$ \\
Compound, rule-compressible &
42 &
$35.0\%$ \\
Compound, intent-compressible &
11 &
$9.2\%$ \\
\hline
\textbf{Total} &
\textbf{120} &
\textbf{$100.0\%$} \\
\hline
\end{tabular}
\caption{Structural distribution of RoboLab-120 under our object-centric event and compression-affordance definitions.}
\label{tab:robolab_task_structure}
\end{table}

\subsection{Grounded Event Extraction}
\label{app:event_extraction}

Our analysis assumes a common grounded event representation $E_\tau$ for each demonstrated task. Here we describe the perception and verbalization pipeline used to recover this representation from demonstration videos. This component provides a shared semantic source for human-specification generation and policy-facing realization; it is not itself the focus of our study.

\textbf{Object-Centric Temporal Representation.}
Given a demonstration $V_\tau=(I_1,\ldots,I_T)$, we convert the video into a sequence of object-centric scene graphs. We use Grounding DINO 1.5 Edge for open-vocabulary object detection \cite{ren2024groundingdino15} and SAM 2 for mask propagation and persistent object tracking \cite{ravi2024sam2}. Grounding DINO is applied to initialize object tracks and to recover objects when tracks are lost, while SAM 2 propagates object masks across intermediate frames. Each frame is represented as a graph $G_t=(\mathcal{V}_t,\mathcal{E}_t)$. Each node stores a persistent object identity together with its category, segmentation mask, bounding box, visual attributes, approximate position, and state. Edges encode spatial and semantic relations between entities, including left/right, front/behind, near, on, and inside. Image-plane relations are computed from object masks, while relative depth is estimated using Depth Anything V2 \cite{yang2024depthanythingv2}. The resulting graph sequence maintains consistent entities and relations across the demonstration.

\textbf{Event Extraction.}
We identify task-relevant events from persistent changes in object states and relations. A candidate change is retained only when the new state or relation remains stable across subsequent frames, reducing transient errors from tracking or segmentation. For the $k$-th transition, we define $\Delta G_k = G_{t_k^{+}} \ominus G_{t_k^{-}},$ where $t_k^{-}$ and $t_k^{+}$ denote representative observations immediately before and after the transition. Consecutive changes involving the same manipulated entity are grouped into a single atomic event. This yields an ordered event sequence $E_\tau=(e_1,\ldots,e_{K_\tau}),$ where each event is represented as $e_k=(A_k,O_k,D_k).$ Here, $A_k$ denotes the manipulation, $O_k$ the manipulated object, and $D_k$ its destination or resulting target state. For example, moving a white mug from the table into a blue bowl produces an event of the form $(\mathrm{place},\mathrm{white\ mug},\mathrm{inside}(\mathrm{blue\ bowl})).$

\textbf{Factual Event Verbalization.}
The structured event representation is used as the semantic source of truth, but directly exposing graph serialization to a language model makes downstream generation unnecessarily dependent on implementation details. We therefore additionally convert each event into a concise factual description. For each transition, Qwen2.5-VL \cite{bai2025qwen25vl} receives the graph change $\Delta G_k$ together with observations immediately before and after the event. It is prompted to identify which entity changed, what changed, and the resulting state or relation. The response is constrained to entities and relations represented in $\Delta G_k$. For example, a transition from $\mathrm{ON}(\mathrm{mug},\mathrm{table})$ to $\mathrm{IN}(\mathrm{mug},\mathrm{bowl})$ may be verbalized as ``The white mug is moved from the table into the blue bowl.'' These descriptions serve only as a natural-language interface to the downstream specification generators. The grounded event record $E_\tau$, rather than the generated description, determines the physical semantics of the task.

\subsection{Human Study and Generation Validation}
\label{app:human_validation}

We conduct a human instruction study for two purposes. First, we use
naturally elicited instructions to characterize how people communicate
physical objectives and to validate that the proposed taxonomy captures
the forms they actually use. Second, we evaluate whether
taxonomy-conditioned generation preserves the intended task semantics
while producing linguistically natural realizations.

\textbf{Task Selection.}
We select 15 tasks from RoboLab spanning single-object manipulation,
multi-object objectives, ordered manipulation sequences, set-level
rules, sorting, reorientation, stacking, and higher-level cleanup
objectives. The selected tasks are:

\begin{center}
\small
\begin{tabular}{p{0.46\linewidth} p{0.46\linewidth}}
\texttt{BananaInBowlTask} &
\texttt{BowlInBinTask} \\

\texttt{YogurtInBowlTask} &
\texttt{WhiteMugInCenterOfTableTask} \\

\texttt{FruitsOnionTask} &
\texttt{RubiksCubeAndBananaTask} \\

\texttt{BananaThenRubiksCubeTask} &
\texttt{RubiksCubeThenBananaTask} \\

\texttt{BlackItemsInBinTask} &
\texttt{ReorientAllMugsTask} \\

\texttt{FoodPackingByColorTask} &
\texttt{BlockStackingSpecifiedOrderTask} \\

\texttt{CleanUpToysTask} &
\texttt{CookingClearPlateTask} \\

\texttt{BananasOutOfBinTask} &
\\
\end{tabular}
\end{center}

The set is chosen to span different numbers of atomic manipulation
events and different opportunities for expressing a compound objective
compactly.

\textbf{Demonstration Generation.}
For each task, we generate a successful demonstration directly from the
RoboLab simulator rather than using a learned policy. The task
definition specifies the relevant target objects and their desired
terminal poses or states. We use the simulator's ground-truth object
mesh models to represent the scene geometry and a motion planner to
generate feasible robot trajectories that move each target object from
its initial pose to the task-defined target pose. For compound tasks,
the procedure is applied to each required manipulation in the order
specified by the task objective. This produces a successful demonstration whose final scene satisfies the benchmark task predicate while avoiding failures or ambiguities
introduced by a learned control policy. Participants observe only the
resulting RGB demonstration video; RoboLab task names, benchmark
instructions, event annotations, and success predicates are not shown.

\textbf{Participants and Ethics.}
We recruit 50 college students between 20 and 27 years old
(31 male and 19 female) for the instruction-elicitation study.
Participants report no prior experience with robotics research or
robot operation. The study involves minimal-risk instruction elicitation and does not
collect sensitive or personally identifying information. Under our
institutional policy, formal IRB review was not required. All
participants provided informed consent before participation, and all
responses were analyzed anonymously.

\textbf{Instruction-Elicitation Procedure.}
Each participant observes demonstrations for all 15 selected tasks,
yielding $50 \times 15 = 750$ human specifications. Task order is
independently randomized for each participant to reduce systematic
ordering effects. For each task, the interface displays the
demonstration video together with a free-form text box. Participants
may replay the video as many times as needed before submitting their
response. No task name, canonical instruction, taxonomy label, or
suggested vocabulary is shown. Participants are told that the robot in the video is acting to satisfy
the objective of a human user. After observing the execution, they are
asked to infer what the user wanted to accomplish and to express how
they would naturally communicate the same objective to another person
who could help complete the task. The prompt is:

\begin{center}
\fbox{%
\begin{minipage}{0.94\linewidth}
\small
\textbf{Prompt shown to participants.}

You will watch a short video of a robot carrying out a task to meet
the intent of a human user. Based on the robot's actions, what do you
think the human user wanted to accomplish?

\medskip

Imagine that you wanted to accomplish the same thing and were asking
another person to help you. How would you naturally communicate what
you want them to do?

\medskip

Write what you would naturally say to that person. There is no required
format or level of detail. You may replay the video as many times as
you need.
\end{minipage}%
}
\end{center}

We deliberately frame the elicitation as human-to-human communication
rather than asking participants to formulate instructions for a robot.
Explicitly addressing a robot may cause speakers to accommodate to
their assumptions about robot capabilities, for example by using more
literal, procedural, or unnaturally explicit commands. Our goal instead
is to capture how people naturally communicate an intended physical
objective, including what they state explicitly and what they expect
another agent to infer from context.

\textbf{Coding Human Specifications.}
We annotate each of the 750 instructions both at the taxonomy level
and at the executable-component level used in the information-gap
analysis. For every task $\tau$, the grounded event sequence 
\[
    E_\tau=(e_1,\ldots,e_{K_\tau}),
    \qquad
    e_k=(A_k,O_k,D_k)
\]
serves as the semantic reference. For each required component
$s\in\{A_k,O_k,D_k\}$, the instruction $u$ receives a label $\sigma_{k,s}(u)\in\{0,1,2\},$ where $\sigma_{k,s}(u)=0$ indicates that the required information is
omitted, $\sigma_{k,s}(u)=1$ indicates that it is expressed but its
grounded value must be recovered from the observed scene or task
context, and $\sigma_{k,s}(u)=2$ indicates that the grounded value is
directly specified in language.

We use two independent model annotators, GPT-5.6 and Claude-Fable-5,
to produce initial annotations for every human specification. Each model
independently assigns the task-content, reference, and composition
labels, together with the component-level $\sigma_{k,s}$ annotations. Two human inspectors subsequently review the model annotations against
the original instruction, demonstration, and grounded event
representation. A label is entered into the final dataset only after
the two human inspectors agree on the annotation. When the initial
model annotations disagree, or when either inspector identifies an
error, the case is reviewed explicitly and resolved by consensus.
Thus, all statistics reported for the human study use the
human-verified consensus annotations rather than either model's
prediction directly.

From the final $\sigma$ annotations, we compute semantic coverage
$\eta_\tau(u)$ and groundedness $\gamma_\tau(u)$ using
Eqs.~\ref{equ:coverage} and~\ref{equ:groundedness}. We separately record
the number $J_\tau(u)$ of distinct manipulation units explicitly
individuated by the instruction and compute the decomposition ratio $ \delta_\tau(u)=\frac{J_\tau(u)}{K_\tau}.$ For example, an instruction that describes three required object
placements in three separate clauses has $\delta_\tau=1$, whereas a
single collective expression covering the same three placements has
$J_\tau=1$ and therefore $\delta_\tau=1/3$. 

\begin{table}[t]
\centering
\footnotesize
\begin{tabular}{
p{0.40\linewidth}
p{0.23\linewidth}
p{0.23\linewidth}
}
\hline
\textbf{Factor} &
\textbf{Agreement (\%)} &
\textbf{Coverage (\%)} \\
\hline
Task content &
$91.3$ &
$98.0$ \\
Reference form &
$96.0$ &
$99.3$ \\
Composition &
$93.3$ &
$98.7$ \\
\hline
Complete specification &
-- &
$96.7$ \\
\hline
\end{tabular}
\caption{Coverage of naturally elicited specifications by the proposed taxonomy and agreement between independent annotators.}
\label{tab:taxonomy_coverage}
\end{table}

\begin{figure*}[t]
    \centering
    \includegraphics[width=\textwidth]{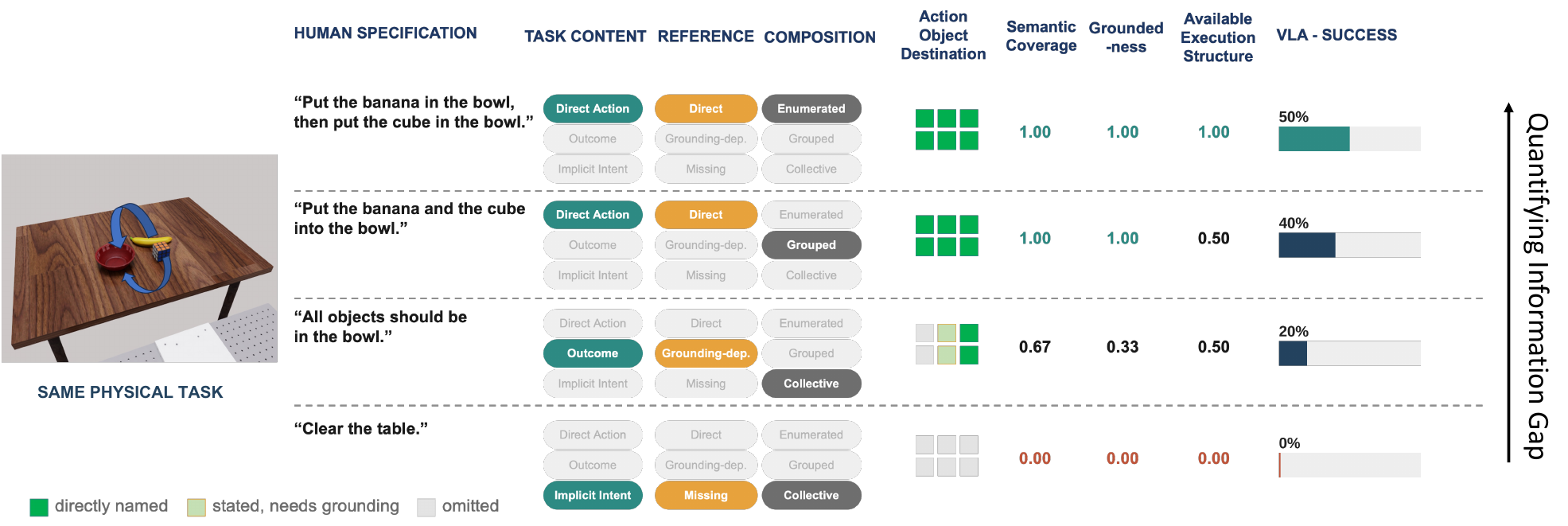}
    \caption{\textbf{From human specifications to measurable information gaps.}
    The same physical objective can be expressed through substantially
    different human specifications. Our taxonomy classifies each
    instruction by its task content, reference form, and composition,
    and maps these choices to the executable information made available
    to the policy: which actions, objects, and destinations are directly
    specified, require grounding, or are omitted. This yields quantitative
    measures of semantic coverage, groundedness, and decomposition ratio. Rather than studying instructions only as lexical variants,
    we can therefore relate VLA performance to the information they carry
    and identify which missing structure the policy can recover and which
    it cannot.}
    \label{fig:information_gap_example}
\end{figure*}

\textbf{Taxonomy Coverage.}
The same consensus procedure is used to assign each specification its
task-content, reference-form, and composition labels. A specification
is considered covered by the taxonomy when all task-relevant aspects
of its objective specification can be represented by these factors
without introducing an additional \emph{Other} category. Table~\ref{tab:taxonomy_coverage} reports annotation agreement before
human adjudication and final taxonomy coverage. The taxonomy covers $96.7\%$ of complete specifications. The remaining
cases are primarily discourse-dependent expressions, such as
``do the same with the rest,'' whose interpretation depends on a
previously established discourse pattern rather than on a recurring
objective-specification form absent from the taxonomy.

\begin{table}[t]
\centering
\footnotesize
\begin{tabular}{
p{0.46\linewidth}
p{0.22\linewidth}
p{0.22\linewidth}
}
\hline
\textbf{Task structure} &
\textbf{Source ID. (\%)} &
\textbf{Correctness (\%)} \\
\hline
Atomic &
$50.8\%$ &
$98.5\%$ \\
Compound, enumerative &
$51.9\%$ &
$97.3\%$ \\
Compound, rule-compressible &
$53.7\%$ &
$95.4\%$ \\
Compound, intent-compressible &
$57.2\%$ &
$90.6\%$ \\
\hline
Overall &
$52.8\%$ &
$96.0\%$ \\
\hline
\end{tabular}
\caption{Human-source identification and semantic correctness of taxonomy-matched generated specifications. Chance source identification is $50\%$.}
\label{tab:human_generation}
\end{table}

\textbf{Matched Human-Likeness Evaluation.}
We evaluate whether taxonomy-conditioned generation produces
linguistically natural specifications, rather than merely satisfying
the requested taxonomy constraints. For each human-written
specification, we use its corresponding task and verified taxonomy
labels, task content, reference form, and composition, to generate
one matched specification. Each resulting pair therefore refers to the
same physical task and instantiates the same taxonomy configuration,
while differing in whether the surface realization was written by a
human or generated by our procedure. Twenty additional participants, disjoint from the instruction-elicitation
study, take part in a blinded source-identification evaluation. On each
trial, participants view the task demonstration together with two
candidate specifications: one human-written and one generated. The
order of the two specifications is randomized independently on every
trial, and the trials themselves are presented in randomized order.
Participants are not told which source produced either specification.

\begin{center}
\fbox{%
\begin{minipage}{0.90\linewidth}
\small
\textbf{Prompt shown to participants.}

Which of these two descriptions do you think was written by a person?
\end{minipage}%
}
\end{center}

We report the fraction of trials on which participants correctly
identify the human-written specification. Chance performance is
$50\%$. Because the two candidates are matched on both the underlying
task and taxonomy configuration, differences in source-identification
accuracy primarily reflect differences in linguistic realization,
rather than differences in the amount or type of task information
communicated.

\textbf{Semantic Fidelity.}
Human-likeness alone does not guarantee that a generated specification
preserves the intended physical objective. We therefore conduct a
separate semantic-fidelity evaluation. A human inspector reviews each
generated specification against the corresponding demonstration and
grounded event sequence $E_\tau$, and determines whether the
specification remains semantically consistent with the demonstrated
task under its assigned taxonomy configuration. The inspector checks whether the generated instruction preserves the
intended task semantics while respecting the requested task-content,
reference, and composition forms. Surface linguistic differences are
not considered errors as long as the resulting specification expresses
the same intended objective at the prescribed level of explicitness. Table~\ref{tab:human_generation} reports source-identification accuracy
and semantic correctness by task structure. Overall source-identification accuracy is $52.8\%$, close to the
$50\%$ chance level, while semantic correctness is $96.0\%$.
Generation remains highly faithful for atomic and enumerative
objectives and becomes somewhat more difficult for increasingly
abstract task structures, particularly intent-compressible objectives.
Together, the taxonomy-coverage, human-likeness, and semantic-fidelity
results support using taxonomy-conditioned generation to construct
controlled human-like specifications for the VLA evaluation.

\subsection{Controlled Specification and Policy-Facing Realization}
\label{app:robot_language}

This appendix describes how we construct the robot-training language
corpus and extract the policy-facing language prior used in
Sec.~\ref{sec:VLA-side}. The goal is not to recover the exact language
seen by a policy during every optimization step, but to characterize the
recurring ways in which grounded manipulation events are expressed in
the robot data on which the evaluated policies were fine-tuned. We use
this prior to construct fully resolved commands whose structure and
surface realization are representative of the policies' robot-training
interface.

\textbf{Corpus Construction.}
All three evaluated policies use checkpoints fine-tuned on DROID:
\(\pi_{0.5}\) with \texttt{pi05\_droid\_jointpos},
\(\pi_0\)-FAST with \texttt{pi0\_fast\_droid\_jointpos}, and
Cosmos3-Edge with \texttt{nvidia/Cosmos3-Edge-Policy-DROID}.
Although the implementations access DROID through different training
pipelines and data formats, their robot-language supervision is derived
from the same underlying dataset. We therefore construct a single
DROID language corpus as a common proxy for the robot-language
distribution shared by the three evaluated checkpoints. We collect the natural-language annotations associated with successful
DROID demonstrations and treat each non-empty annotation as a corpus
instance. When multiple annotations are available for the same
demonstration, they are retained as separate instances. We do not
deduplicate paraphrases, since repeated linguistic forms provide
evidence about their prevalence in the training distribution. We apply
only minimal text normalization, such as removing formatting artifacts
and normalizing whitespace, before extracting command structure. The corpus is constructed independently of our evaluation language.
In particular, RoboLab benchmark instructions, human-study
specifications, and taxonomy-conditioned specifications generated for
our experiments are excluded. Thus, the policy-facing realizations are
derived only from robot-training language and do not use language from
the evaluation tasks themselves.

\textbf{Manipulation-Clause Extraction.}
We analyze the corpus at the level of manipulation clauses corresponding
to the event representation used throughout the paper. Each clause is
mapped, when possible, onto the semantic structure
\((A,O,D)\), where \(A\) denotes the manipulation, \(O\) the manipulated
object, and \(D\) the destination or resulting target state. For
instructions that explicitly contain multiple manipulation clauses, we
treat the clauses separately when their event boundaries are
unambiguous. This produces a collection of language realizations
aligned with the same event granularity as
\(e_k=(A_k,O_k,D_k)\). This processing is used only to extract linguistic regularities. It
does not determine the event sequence \(E_\tau\) of a RoboLab task,
which is obtained independently from the grounded event extraction
pipeline described in Appendix~\ref{app:event_extraction}.

\textbf{Template Normalization.}
To separate recurring command structure from task-specific entities, we
replace spans referring to manipulated objects and destinations with
typed placeholders. For example, $ \text{``Put the red mug inside the blue bowl''}$ is normalized to $\text{``Put [OBJECT] inside [DESTINATION].''}$ Instructions with the same normalized structure are grouped under a
shared template \(t\). Because the corpus is shared across the three
policies, we estimate a single empirical template distribution
\[
    p(t)
    =
    \frac{N(t)}
         {\sum_{t'} N(t')},
\]
where \(N(t)\) is the number of corpus instances realizing template
\(t\). We preserve these empirical frequencies rather than treating
all observed templates as equally representative of the training
language. Templates retain the linguistic frame of the command, including its
action and relational wording. Consequently, surface forms such as
``put [OBJECT] in [DESTINATION]'' and
``place [OBJECT] inside [DESTINATION]'' may correspond to different
templates even when they express the same grounded event. This allows
the corpus to capture both structural and lexical regularities of
robot-training instructions.

\textbf{Lexical Realization.}
In addition to full-template frequency, we record the lexical forms
used to express equivalent semantic components. For actions, this
includes alternatives such as ``pick up,'' ``grasp,'' or ``lift.'' For
destinations and target states, it includes alternative relational
expressions such as ``in,'' ``inside,'' ``onto,'' or ``on top of,''
when they correspond to the same grounded relation. For a semantic component, we estimate the empirical frequency of
each lexical realization \(\ell\) among compatible corpus instances.
These statistics allow a grounded event to be expressed using language
that is actually represented in the robot-training corpus rather than
through unconstrained paraphrasing.

\textbf{Reference Forms.}
We separately characterize how manipulated objects and destinations are
referred to in the corpus. Reference expressions are grouped according
to the information used to identify the grounded entity, including
category names, attribute-qualified category names, attribute-based
expressions, and relational descriptions. For example, depending on
the available scene information, a grounded entity may be described as
``the mug,'' ``the yellow mug,'' ``the yellow object,'' or
``the object beside the plate.'' For each reference form, we retain both its structure and its empirical
frequency in the DROID corpus. This distinction is important because
two commands can share the same action semantics and syntactic
structure while exposing different amounts of information about the
referent.

\textbf{Policy-Facing Realization Generation.}
Given a grounded event $e_k=(A_k,O_k,D_k),$ we first identify the subset of corpus templates semantically compatible with the event's action and target relation. A compatible template is
then selected according to its empirical corpus frequency and
instantiated with references to \(O_k\) and \(D_k\). Lexical choices
and reference forms are likewise restricted to realizations observed
for compatible semantic components in the DROID corpus. When a
reference form requires an attribute or spatial relation, the required
information is obtained from the grounded scene representation
associated with the task.

The resulting command $r_k = \mathcal{R}(e_k)$ therefore expresses the complete grounded event while remaining close
to the language interface represented in robot-training data. For a
compound task $E_\tau=(e_1,\ldots,e_{K_\tau}),$ the procedure is applied independently to each event, yielding the
atomic policy-facing sequence $R_\tau=(r_1,\ldots,r_{K_\tau}).$ Each \(r_k\) explicitly identifies the required manipulation,
manipulated object, and destination or target state. The sequence
\(R_\tau\) is therefore a fully resolved realization of the physical
objective, rather than an alternative interpretation of it.

For the controlled communication-gap experiments, the selected
policy-facing realizations are fixed across matched conditions. Thus,
when human-like specifications are compared with the fully resolved
reference, differences in behavior cannot be attributed to changes in
the underlying event semantics or to independently varied command
wording. The same atomic realizations are also used in the residual
execution-gap experiments, where we vary only how the resolved events
are organized and invoked.
\subsection{Execution Protocol and Statistical Analysis}
\label{app:evaluation_protocol}

This section provides the execution details for the information-gap
and residual execution-gap evaluations. Within each task, all compared
conditions use the same physical objective, initial scene configuration,
success criteria, and task-execution budget.

\textbf{Simulation and Policy Observations.}
Experiments are run in RoboLab v0.3.1 using Isaac Sim 5.0.0 and
Isaac Lab 2.2.0. Physics is simulated at 120 Hz, with a control
decimation of 8, giving a policy control frequency of 15 Hz.
The policy observes two RGB views: an over-the-shoulder left camera
and a wrist-mounted camera. Both views are resized to $224\times224$
before being provided to the policy, together with the robot
proprioceptive state.

\textbf{Behavioral Metrics.}
We report task success and normalized partial progress.
\emph{Success} is determined by the RoboLab success predicate for the
corresponding task. \emph{Progress} measures the fraction of the task
objective completed when the full success predicate is not satisfied.
Both metrics are used for the information-gap and residual
execution-gap analyses, with
$M\in\{\mathrm{Success},\mathrm{Progress}\}$.

\textbf{Execution Budget.}
Execution budgets are measured in policy-control steps at 15 Hz.
For a task with benchmark episode horizon $T_\tau$ seconds, the total
task budget is $ B_\tau = 15 T_\tau$
control steps. The evaluated compound tasks use horizons of
$60$, $90$, $120$, $180$, or $300$ seconds, corresponding to total
budgets of $900$, $1350$, $1800$, $2700$, or $4500$ task steps,
respectively. All presentation conditions receive the same total task budget. Under
sequential execution, a task with $K_\tau$ atomic events is assigned
absolute event deadlines $b_k =
    \left\lceil
    \frac{k B_\tau}{K_\tau}
    \right\rceil,
    \qquad k=1,\ldots,K_\tau.$
Thus, the nominal allocation is $B_\tau/K_\tau$ steps per event, but
unused time from an early-completed event remains available to later
events. In the evaluated tasks, nominal per-event allocations are
typically $450$ steps ($30$ s), with longer allocations for tasks
whose benchmark horizon is correspondingly larger. If an event has
not completed by its deadline, it is marked as timed out and execution
advances to the next event.

\textbf{Information-Gap Execution.}
For each human-like specification $u$ and the fully resolved reference
$r_\tau^{\mathrm{full}}$, the complete specification is supplied once
at the beginning of the rollout and remains fixed throughout execution.
The policy acts until the benchmark success condition is satisfied or
the total task budget is exhausted. The reference $r_\tau^{\mathrm{full}}$ uses
Action--Direct--Enumerated semantics: every required action, object,
destination, and atomic event in $E_\tau$ is explicitly specified.
This provides the matched reference used to measure the behavioral
cost of unresolved task information.

\textbf{Residual Execution-Gap Conditions.}
To study execution after the information gap has been removed, we hold
the fully grounded atomic command sequence $R_\tau=(r_1,\ldots,r_{K_\tau})$ fixed and vary only how it is presented and invoked. In the \emph{one-prompt} condition, all atomic commands are provided
together at the beginning of the rollout. In the \emph{sequential}
condition, the policy receives only $r_1$ initially and advances to
$r_{k+1}$ when the completion predicate for event $e_k$ is satisfied
or its deadline is reached. When the command changes, the policy's
cached action chunk is cleared so that the next action is generated
from the new command.

\textbf{Per-Event Completion Predicates.}
Sequential execution requires a completion criterion for each atomic
event. We derive these predicates from RoboLab's task definition and
subtask structure rather than introducing separate manually specified
success criteria. Each policy-facing command $r_k$ is associated with
the corresponding RoboLab stage or subtask group whose completion
represents event $e_k$. During execution, the relevant RoboLab subtask state is queried
read-only after each environment step. An event is considered complete
when its associated stage or object group is marked complete by the
benchmark. For repeated, visually interchangeable objects, completion
is defined by the required count of completed object groups rather than
by a particular object identity. Event completion is latched for the
purpose of advancing through the command sequence.

\textbf{Home-Reset Intervention.}
We additionally evaluate sequential execution with a home reset between
events. After event $e_k$ completes or reaches its deadline, and before
$r_{k+1}$ is issued, the gripper is opened for 6 control steps and the
arm is moved back to the environment's default home configuration by
a 45-step linear interpolation in joint space. The robot then holds
the home configuration for up to 10 additional steps, terminating the
settling phase early when the maximum joint-position error falls below
$0.02$ rad. There is no reset before the first command or after the
final command, and the simulator scene itself is not reset; object
states produced by preceding events remain in the environment.

Reset motion does not consume the task-step budget used for policy
execution. It does, however, consume simulated time. We therefore
increase the environment time limit by up to $61(K_\tau-1)$ environment steps in the home-reset condition, corresponding to at most
$6+45+10=61$ reset steps at each event boundary. Consequently, the
one-prompt, sequential, and sequential-plus-home-reset conditions all
receive the same policy-controlled task budget.

\textbf{Matched Evaluation.}
Each task--instruction condition is evaluated over 10 rollouts.
Comparisons within a task use the same physical objective, initial
scene distribution, evaluation criteria, and task budget. The
information-gap experiments therefore vary only the information
available in the specification, while the residual execution-gap
experiments hold the resolved event sequence fixed and vary only its
presentation and invocation state.

\textbf{Aggregation and Confidence Intervals.}
The task is the unit of statistical analysis. For each
task--condition pair, we first average outcomes over the 10 rollout
trials and, where applicable, over surface realizations. Aggregate
success and progress are then computed over these task-level means. We report 95\% confidence intervals using a nonparametric bootstrap
over tasks with 10,000 resamples. For matched comparisons, including
the information-gap quantities
$\Delta^{\mathrm{info}}_M$, we first compute the within-task difference
between the compared conditions and bootstrap these paired task-level
differences. Thus, repeated rollouts of the same task are not treated
as independent samples.

\textbf{Cost per Unresolved Unit.}
To estimate the behavioral cost of different forms of unresolved task
structure in Fig.~\ref{fig:information_gap_behavior}, we define, for each
task--specification pair $i$, the number of missing semantic components
$m_i$, grounding-dependent components $g_i$, and undecomposed events
$d_i$:
\[
m_i =
\sum_{k,s}\mathbbm{1}[\sigma_{k,s}=0],
\qquad
g_i =
\sum_{k,s}\mathbbm{1}[\sigma_{k,s}=1],
\qquad
d_i = K_{\tau_i}-J_{\tau_i}.
\]
For each policy separately, we fit
\[
\Delta^{\mathrm{info}}_{\mathrm{Success},i}
=
\beta_{\mathrm{miss}}m_i
+
\beta_{\mathrm{ground}}g_i
+
\beta_{\mathrm{decomp}}d_i
+
\epsilon_i.
\]
The regression is fit without an intercept because the fully resolved
reference has zero unresolved units and, by construction, zero matched
information gap. The coefficients therefore give the estimated
percentage-point success cost per missing semantic component,
grounding-dependent component, and undecomposed event, respectively.
Uncertainty is estimated by refitting the regression under the same
task-level bootstrap procedure.

\textbf{Association with Available Structure.}
For Fig.~\ref{fig:information_gap_behavior}, we additionally measure how strongly
the amount of available task structure is associated with the
behavioral information gap as task complexity increases. Within each
complexity group ($K_\tau=1$, $K_\tau=2$, and $K_\tau\ge3$), we compute
Spearman's rank correlation between
$\Delta^{\mathrm{info}}_{\mathrm{Success}}$ and each of semantic
coverage $\eta_\tau$, groundedness $\gamma_\tau$, and decomposition ratio
$\delta_\tau$. Decomposition ratio is omitted for single-event tasks, for
which $\delta_\tau=1$ by definition. More negative correlation
indicates that greater availability of the corresponding executable
structure is associated with a smaller success gap.

\textbf{Reference for Grouped and Collective Conditions}
Grouped and collective specifications exist only for compound tasks, and we evaluate them on 54 and 50 of the 56 compound tasks, respectively. For these columns in Table~\ref{tab:information_gap_main} and Tables~\ref{tab:pi-0.5-main-table}--\ref{tab:cosmos-main-table}, we compute $\Delta^{\mathrm{info}}_{\mathrm{Success}}$ against the Action--Direct--Enumerated reference averaged over all 56 compound tasks, which is the one-prompt condition in Fig.~\ref{fig:observation_alignment_gap} (23.5\% for $\pi_{0.5}$, 8.2\% for $\pi_0$-FAST, and 20.1\% for Cosmos3-Edge).This departs from the per-subset matching used for the enumerated column.
\subsection{Additional Models}
\label{app:additional_models}

\begin{table*}[t]
\centering
\scriptsize
\setlength{\tabcolsep}{4pt}
\begin{adjustbox}{max width=\linewidth}
\begin{tabular}{lllccccccc}
\hline
\multicolumn{3}{c}{\textbf{Taxonomy configuration}} &
\multicolumn{3}{c}{\textbf{Specification structure}} &
\multicolumn{3}{c}{\textbf{VLA behavior}} &
\\
\textbf{Content} &
\textbf{Reference} &
\textbf{Composition} &
$\boldsymbol{\eta_\tau}$ &
$\boldsymbol{\gamma_\tau}$ &
$\boldsymbol{\bar{\delta}}$ &
\textbf{Success (\%)} &
\textbf{Progress} &
$\boldsymbol{\Delta_{\mathrm{Success}}^{\mathrm{info}}}$ &
$\boldsymbol{N}$ \\
\hline
Action & Direct & Enumerated & 1.00 & 1.00 & 1.00 $\pm$ 0.00 & 38.0 $\pm$ 5.4 & 0.51 $\pm$ 0.05 & \cellcolor{gapred!0}0.0 (ref.) & 120 \\
 &  & Grouped & 1.00 & 1.00 & 0.47 $\pm$ 0.08 & 15.9 $\pm$ 6.0 & 0.39 $\pm$ 0.06 & \cellcolor{gapred!23}\textbf{7.6 $\pm$ 3.1} & 54 \\
 &  & Collective & 1.00 & 1.00 & 0.41 $\pm$ 0.04 & 13.1 $\pm$ 5.8 & 0.35 $\pm$ 0.06 & \cellcolor{gapred!31}\textbf{10.4 $\pm$ 3.5} & 50 \\
\cline{2-10}
 & Grounding-dep. & Enumerated & 1.00 & 0.33 & 1.00 $\pm$ 0.00 & 35.2 $\pm$ 5.3 & 0.48 $\pm$ 0.05 & \cellcolor{gapred!8}\textbf{2.8 $\pm$ 1.8} & 120 \\
 &  & Grouped & 1.00 & 0.33 & 0.47 $\pm$ 0.08 & 13.5 $\pm$ 5.6 & 0.35 $\pm$ 0.05 & \cellcolor{gapred!30}10.0 $\pm$ 3.3 & 54 \\
 &  & Collective & 1.00 & 0.33 & 0.41 $\pm$ 0.04 & 12.4 $\pm$ 5.7 & 0.34 $\pm$ 0.05 & \cellcolor{gapred!33}11.1 $\pm$ 3.6 & 50 \\
\cline{2-10}
 & Missing & Enumerated & 0.33 & 0.33 & 1.00 $\pm$ 0.00 & 26.9 $\pm$ 4.9 & 0.38 $\pm$ 0.05 & \cellcolor{gapred!33}\textbf{11.1 $\pm$ 2.3} & 120 \\
 &  & Grouped & 0.33 & 0.33 & 0.47 $\pm$ 0.08 & 7.1 $\pm$ 4.2 & 0.24 $\pm$ 0.04 & \cellcolor{gapred!49}16.4 $\pm$ 3.9 & 54 \\
 &  & Collective & 0.33 & 0.33 & 0.41 $\pm$ 0.04 & 6.5 $\pm$ 4.2 & 0.23 $\pm$ 0.04 & \cellcolor{gapred!51}17.0 $\pm$ 4.1 & 50 \\
\hline
Outcome & Direct & Enumerated & 0.67 & 0.67 & 1.00 $\pm$ 0.00 & 33.5 $\pm$ 5.2 & 0.46 $\pm$ 0.05 & \cellcolor{gapred!14}\textbf{4.5 $\pm$ 1.9} & 120 \\
 &  & Grouped & 0.67 & 0.67 & 0.47 $\pm$ 0.08 & 11.8 $\pm$ 5.3 & 0.33 $\pm$ 0.05 & \cellcolor{gapred!35}11.7 $\pm$ 3.5 & 54 \\
 &  & Collective & 0.67 & 0.67 & 0.41 $\pm$ 0.04 & 9.7 $\pm$ 5.1 & 0.29 $\pm$ 0.05 & \cellcolor{gapred!41}13.8 $\pm$ 3.8 & 50 \\
\cline{2-10}
 & Grounding-dep. & Enumerated & 0.67 & 0.00 & 1.00 $\pm$ 0.00 & 31.6 $\pm$ 5.2 & 0.44 $\pm$ 0.05 & \cellcolor{gapred!19}6.4 $\pm$ 2.0 & 120 \\
 &  & Grouped & 0.67 & 0.00 & 0.47 $\pm$ 0.08 & 10.0 $\pm$ 5.0 & 0.30 $\pm$ 0.05 & \cellcolor{gapred!41}13.5 $\pm$ 3.7 & 54 \\
 &  & Collective & 0.67 & 0.00 & 0.41 $\pm$ 0.04 & 9.2 $\pm$ 5.0 & 0.28 $\pm$ 0.05 & \cellcolor{gapred!43}14.3 $\pm$ 3.9 & 50 \\
\cline{2-10}
 & Missing & Enumerated & 0.00 & 0.00 & 1.00 $\pm$ 0.00 & 24.6 $\pm$ 4.8 & 0.35 $\pm$ 0.05 & \cellcolor{gapred!40}13.4 $\pm$ 2.4 & 120 \\
 &  & Grouped & 0.00 & 0.00 & 0.47 $\pm$ 0.08 & 5.3 $\pm$ 3.7 & 0.20 $\pm$ 0.04 & \cellcolor{gapred!55}18.2 $\pm$ 4.1 & 54 \\
 &  & Collective & 0.00 & 0.00 & 0.41 $\pm$ 0.04 & 4.9 $\pm$ 3.7 & 0.19 $\pm$ 0.04 & \cellcolor{gapred!56}18.6 $\pm$ 4.3 & 50 \\
\hline
Intent & Direct & Enumerated & 0.33 & 0.33 & 1.00 $\pm$ 0.00 & 22.8 $\pm$ 4.7 & 0.33 $\pm$ 0.04 & \cellcolor{gapred!46}\textbf{15.2 $\pm$ 2.6} & 120 \\
 &  & Grouped & 0.33 & 0.33 & 0.47 $\pm$ 0.08 & 5.3 $\pm$ 3.7 & 0.20 $\pm$ 0.04 & \cellcolor{gapred!55}18.2 $\pm$ 4.1 & 54 \\
 &  & Collective & 0.33 & 0.33 & 0.41 $\pm$ 0.04 & 4.3 $\pm$ 3.5 & 0.18 $\pm$ 0.03 & \cellcolor{gapred!58}19.2 $\pm$ 4.4 & 50 \\
\cline{2-10}
 & Grounding-dep. & Enumerated & 0.33 & 0.00 & 1.00 $\pm$ 0.00 & 22.1 $\pm$ 4.6 & 0.32 $\pm$ 0.04 & \cellcolor{gapred!48}15.9 $\pm$ 2.6 & 120 \\
 &  & Grouped & 0.33 & 0.00 & 0.47 $\pm$ 0.08 & 4.7 $\pm$ 3.5 & 0.19 $\pm$ 0.03 & \cellcolor{gapred!56}18.8 $\pm$ 4.2 & 54 \\
 &  & Collective & 0.33 & 0.00 & 0.41 $\pm$ 0.04 & 4.3 $\pm$ 3.5 & 0.18 $\pm$ 0.03 & \cellcolor{gapred!58}19.2 $\pm$ 4.4 & 50 \\
\cline{2-10}
 & Missing & Enumerated & 0.00 & 0.00 & 1.00 $\pm$ 0.00 & 18.8 $\pm$ 4.3 & 0.27 $\pm$ 0.04 & \cellcolor{gapred!58}19.3 $\pm$ 2.8 & 120 \\
 &  & Grouped & 0.00 & 0.00 & 0.47 $\pm$ 0.08 & 3.2 $\pm$ 2.9 & 0.15 $\pm$ 0.03 & \cellcolor{gapred!61}20.3 $\pm$ 4.3 & 54 \\
 &  & Collective & 0.00 & 0.00 & 0.41 $\pm$ 0.04 & 2.9 $\pm$ 2.9 & 0.14 $\pm$ 0.03 & \cellcolor{gapred!62}20.6 $\pm$ 4.5 & 50 \\
\hline
 \end{tabular}
\end{adjustbox}
\caption{$\pi_{0.5}$ performance across the full factorial space of
taxonomy-controlled specifications for all 120 tasks.
Enumerated composition applies to all 120 tasks, while grouped and collective
forms are evaluated on 54 and 50 eligible tasks, respectively.
Results are first averaged over 10 rollouts and multiple surface realizations
for each task--configuration pair, and then across eligible tasks.
Behavioral results are reported as mean $\pm$ 95\% confidence-interval
half-width, computed over task-level means.
$\eta_\tau$, $\gamma_\tau$, and $\delta_\tau$ denote semantic coverage,
groundedness, and decomposition ratio; the first two are fixed by
the content and reference levels, while $\bar{\delta}$ is the mean of
$\delta_\tau$ over eligible tasks.
$\Delta_{\mathrm{Success}}^{\mathrm{info}}$ is a matched task-level difference
relative to $r_\tau^{\mathrm{full}}$ over the same eligible task subset (for grouped and collective rows, the reference is averaged over all 56 compound tasks);
shading scales with its magnitude, and bold marks configurations that vary
a single factor from the reference.
}
\label{tab:pi-0.5-main-table}
\vspace{-15pt}
\end{table*}

\begin{table*}[t]
\centering
\scriptsize
\setlength{\tabcolsep}{4pt}
\begin{adjustbox}{max width=\linewidth}
\begin{tabular}{lllccccccc}
\hline
\multicolumn{3}{c}{\textbf{Taxonomy configuration}} &
\multicolumn{3}{c}{\textbf{Specification structure}} &
\multicolumn{3}{c}{\textbf{VLA behavior}} &
\\
\textbf{Content} &
\textbf{Reference} &
\textbf{Composition} &
$\boldsymbol{\eta_\tau}$ &
$\boldsymbol{\gamma_\tau}$ &
$\boldsymbol{\bar{\delta}}$ &
\textbf{Success (\%)} &
\textbf{Progress} &
$\boldsymbol{\Delta_{\mathrm{Success}}^{\mathrm{info}}}$ &
$\boldsymbol{N}$ \\
\hline
Action & Direct & Enumerated & 1.00 & 1.00 & 1.00 $\pm$ 0.00 & 22.0 $\pm$ 4.6 & 0.32 $\pm$ 0.04 & \cellcolor{gapred!0}0.0 (ref.) & 120 \\
 &  & Grouped & 1.00 & 1.00 & 0.47 $\pm$ 0.08 & 3.2 $\pm$ 2.9 & 0.18 $\pm$ 0.03 & \cellcolor{gapred!15}\textbf{5.0 $\pm$ 2.0} & 54 \\
 &  & Collective & 1.00 & 1.00 & 0.41 $\pm$ 0.04 & 2.0 $\pm$ 2.4 & 0.16 $\pm$ 0.02 & \cellcolor{gapred!19}\textbf{6.2 $\pm$ 2.2} & 50 \\
\cline{2-10}
 & Grounding-dep. & Enumerated & 1.00 & 0.33 & 1.00 $\pm$ 0.00 & 18.9 $\pm$ 4.3 & 0.28 $\pm$ 0.04 & \cellcolor{gapred!9}\textbf{3.1 $\pm$ 1.6} & 120 \\
 &  & Grouped & 1.00 & 0.33 & 0.47 $\pm$ 0.08 & 1.9 $\pm$ 2.3 & 0.16 $\pm$ 0.02 & \cellcolor{gapred!19}6.3 $\pm$ 2.1 & 54 \\
 &  & Collective & 1.00 & 0.33 & 0.41 $\pm$ 0.04 & 1.7 $\pm$ 2.2 & 0.15 $\pm$ 0.02 & \cellcolor{gapred!20}6.5 $\pm$ 2.2 & 50 \\
\cline{2-10}
 & Missing & Enumerated & 0.33 & 0.33 & 1.00 $\pm$ 0.00 & 14.1 $\pm$ 3.9 & 0.22 $\pm$ 0.04 & \cellcolor{gapred!24}\textbf{7.9 $\pm$ 1.9} & 120 \\
 &  & Grouped & 0.33 & 0.33 & 0.47 $\pm$ 0.08 & 0.6 $\pm$ 1.3 & 0.13 $\pm$ 0.01 & \cellcolor{gapred!23}7.6 $\pm$ 2.2 & 54 \\
 &  & Collective & 0.33 & 0.33 & 0.41 $\pm$ 0.04 & 0.4 $\pm$ 1.1 & 0.13 $\pm$ 0.01 & \cellcolor{gapred!23}7.8 $\pm$ 2.3 & 50 \\
\hline
Outcome & Direct & Enumerated & 0.67 & 0.67 & 1.00 $\pm$ 0.00 & 18.7 $\pm$ 4.3 & 0.28 $\pm$ 0.04 & \cellcolor{gapred!10}\textbf{3.3 $\pm$ 1.6} & 120 \\
 &  & Grouped & 0.67 & 0.67 & 0.47 $\pm$ 0.08 & 1.7 $\pm$ 2.1 & 0.15 $\pm$ 0.02 & \cellcolor{gapred!20}6.5 $\pm$ 2.1 & 54 \\
 &  & Collective & 0.67 & 0.67 & 0.41 $\pm$ 0.04 & 1.0 $\pm$ 1.7 & 0.14 $\pm$ 0.02 & \cellcolor{gapred!22}7.2 $\pm$ 2.3 & 50 \\
\cline{2-10}
 & Grounding-dep. & Enumerated & 0.67 & 0.00 & 1.00 $\pm$ 0.00 & 16.7 $\pm$ 4.1 & 0.25 $\pm$ 0.04 & \cellcolor{gapred!16}5.3 $\pm$ 1.7 & 120 \\
 &  & Grouped & 0.67 & 0.00 & 0.47 $\pm$ 0.08 & 1.0 $\pm$ 1.6 & 0.14 $\pm$ 0.02 & \cellcolor{gapred!22}7.2 $\pm$ 2.2 & 54 \\
 &  & Collective & 0.67 & 0.00 & 0.41 $\pm$ 0.04 & 0.9 $\pm$ 1.6 & 0.14 $\pm$ 0.02 & \cellcolor{gapred!22}7.3 $\pm$ 2.3 & 50 \\
\cline{2-10}
 & Missing & Enumerated & 0.00 & 0.00 & 1.00 $\pm$ 0.00 & 12.5 $\pm$ 3.7 & 0.20 $\pm$ 0.03 & \cellcolor{gapred!28}9.5 $\pm$ 2.0 & 120 \\
 &  & Grouped & 0.00 & 0.00 & 0.47 $\pm$ 0.08 & 0.3 $\pm$ 0.9 & 0.13 $\pm$ 0.01 & \cellcolor{gapred!24}7.9 $\pm$ 2.3 & 54 \\
 &  & Collective & 0.00 & 0.00 & 0.41 $\pm$ 0.04 & 0.2 $\pm$ 0.8 & 0.13 $\pm$ 0.01 & \cellcolor{gapred!24}8.0 $\pm$ 2.4 & 50 \\
\hline
Intent & Direct & Enumerated & 0.33 & 0.33 & 1.00 $\pm$ 0.00 & 11.4 $\pm$ 3.5 & 0.19 $\pm$ 0.03 & \cellcolor{gapred!32}\textbf{10.6 $\pm$ 2.0} & 120 \\
 &  & Grouped & 0.33 & 0.33 & 0.47 $\pm$ 0.08 & 0.3 $\pm$ 0.9 & 0.13 $\pm$ 0.01 & \cellcolor{gapred!24}7.9 $\pm$ 2.3 & 54 \\
 &  & Collective & 0.33 & 0.33 & 0.41 $\pm$ 0.04 & 0.2 $\pm$ 0.8 & 0.13 $\pm$ 0.01 & \cellcolor{gapred!24}8.0 $\pm$ 2.4 & 50 \\
\cline{2-10}
 & Grounding-dep. & Enumerated & 0.33 & 0.00 & 1.00 $\pm$ 0.00 & 10.7 $\pm$ 3.4 & 0.18 $\pm$ 0.03 & \cellcolor{gapred!34}11.3 $\pm$ 2.1 & 120 \\
 &  & Grouped & 0.33 & 0.00 & 0.47 $\pm$ 0.08 & 0.2 $\pm$ 0.7 & 0.13 $\pm$ 0.01 & \cellcolor{gapred!24}8.0 $\pm$ 2.3 & 54 \\
 &  & Collective & 0.33 & 0.00 & 0.41 $\pm$ 0.04 & 0.2 $\pm$ 0.8 & 0.13 $\pm$ 0.01 & \cellcolor{gapred!24}8.0 $\pm$ 2.4 & 50 \\
\cline{2-10}
 & Missing & Enumerated & 0.00 & 0.00 & 1.00 $\pm$ 0.00 & 8.8 $\pm$ 3.1 & 0.15 $\pm$ 0.03 & \cellcolor{gapred!40}13.2 $\pm$ 2.2 & 120 \\
 &  & Grouped & 0.00 & 0.00 & 0.47 $\pm$ 0.08 & 0.1 $\pm$ 0.5 & 0.12 $\pm$ 0.01 & \cellcolor{gapred!24}8.1 $\pm$ 2.3 & 54 \\
 &  & Collective & 0.00 & 0.00 & 0.41 $\pm$ 0.04 & 0.1 $\pm$ 0.5 & 0.12 $\pm$ 0.01 & \cellcolor{gapred!24}8.1 $\pm$ 2.4 & 50 \\
\hline
\end{tabular}
\end{adjustbox}
\caption{$\pi_{0}$-FAST performance across the full factorial space of
taxonomy-controlled specifications for all 120 tasks.
Enumerated composition applies to all 120 tasks, while grouped and collective
forms are evaluated on 54 and 50 eligible tasks, respectively.
Results are first averaged over 10 rollouts and multiple surface realizations
for each task--configuration pair, and then across eligible tasks.
Behavioral results are reported as mean $\pm$ 95\% confidence-interval
half-width, computed over task-level means.
$\eta_\tau$, $\gamma_\tau$, and $\delta_\tau$ denote semantic coverage,
groundedness, and decomposition ratio; the first two are fixed by
the content and reference levels, while $\bar{\delta}$ is the mean of
$\delta_\tau$ over eligible tasks.
$\Delta_{\mathrm{Success}}^{\mathrm{info}}$ is a matched task-level difference
relative to $r_\tau^{\mathrm{full}}$ over the same eligible task subset (for grouped and collective rows, the reference is averaged over all 56 compound tasks);
shading scales with its magnitude, and bold marks configurations that vary
a single factor from the reference.
}
\label{tab:pi0fast-main-table}
\vspace{-15pt}
\end{table*}

\begin{table*}[t]
\centering
\scriptsize
\setlength{\tabcolsep}{4pt}
\begin{adjustbox}{max width=\linewidth}
\begin{tabular}{lllccccccc}
\hline
\multicolumn{3}{c}{\textbf{Taxonomy configuration}} &
\multicolumn{3}{c}{\textbf{Specification structure}} &
\multicolumn{3}{c}{\textbf{WAM behavior}} &
\\
\textbf{Content} &
\textbf{Reference} &
\textbf{Composition} &
$\boldsymbol{\eta_\tau}$ &
$\boldsymbol{\gamma_\tau}$ &
$\boldsymbol{\bar{\delta}}$ &
\textbf{Success (\%)} &
\textbf{Progress} &
$\boldsymbol{\Delta_{\mathrm{Success}}^{\mathrm{info}}}$ &
$\boldsymbol{N}$ \\
\hline
Action & Direct & Enumerated & 1.00 & 1.00 & 1.00 $\pm$ 0.00 & 33.0 $\pm$ 5.2 & 0.45 $\pm$ 0.05 & \cellcolor{gapred!0}0.0 (ref.) & 120 \\
 &  & Grouped & 1.00 & 1.00 & 0.47 $\pm$ 0.08 & 13.1 $\pm$ 5.6 & 0.31 $\pm$ 0.05 & \cellcolor{gapred!21}\textbf{7.0 $\pm$ 2.9} & 54 \\
 &  & Collective & 1.00 & 1.00 & 0.41 $\pm$ 0.04 & 10.6 $\pm$ 5.3 & 0.27 $\pm$ 0.05 & \cellcolor{gapred!28}\textbf{9.5 $\pm$ 3.3} & 50 \\
\cline{2-10}
 & Grounding-dep. & Enumerated & 1.00 & 0.33 & 1.00 $\pm$ 0.00 & 31.0 $\pm$ 5.1 & 0.43 $\pm$ 0.05 & \cellcolor{gapred!6}\textbf{2.0 $\pm$ 1.7} & 120 \\
 &  & Grouped & 1.00 & 0.33 & 0.47 $\pm$ 0.08 & 11.4 $\pm$ 5.3 & 0.28 $\pm$ 0.05 & \cellcolor{gapred!26}8.7 $\pm$ 3.1 & 54 \\
 &  & Collective & 1.00 & 0.33 & 0.41 $\pm$ 0.04 & 10.1 $\pm$ 5.2 & 0.26 $\pm$ 0.05 & \cellcolor{gapred!30}10.0 $\pm$ 3.3 & 50 \\
\cline{2-10}
 & Missing & Enumerated & 0.33 & 0.33 & 1.00 $\pm$ 0.00 & 23.1 $\pm$ 4.7 & 0.33 $\pm$ 0.04 & \cellcolor{gapred!30}\textbf{9.9 $\pm$ 2.2} & 120 \\
 &  & Grouped & 0.33 & 0.33 & 0.47 $\pm$ 0.08 & 5.6 $\pm$ 3.8 & 0.17 $\pm$ 0.04 & \cellcolor{gapred!44}14.5 $\pm$ 3.6 & 54 \\
 &  & Collective & 0.33 & 0.33 & 0.41 $\pm$ 0.04 & 5.1 $\pm$ 3.8 & 0.16 $\pm$ 0.04 & \cellcolor{gapred!45}15.0 $\pm$ 3.8 & 50 \\
\hline
Outcome & Direct & Enumerated & 0.67 & 0.67 & 1.00 $\pm$ 0.00 & 29.0 $\pm$ 5.0 & 0.40 $\pm$ 0.05 & \cellcolor{gapred!12}\textbf{4.0 $\pm$ 1.8} & 120 \\
 &  & Grouped & 0.67 & 0.67 & 0.47 $\pm$ 0.08 & 9.6 $\pm$ 4.9 & 0.25 $\pm$ 0.05 & \cellcolor{gapred!32}10.5 $\pm$ 3.2 & 54 \\
 &  & Collective & 0.67 & 0.67 & 0.41 $\pm$ 0.04 & 7.7 $\pm$ 4.6 & 0.21 $\pm$ 0.04 & \cellcolor{gapred!37}12.4 $\pm$ 3.5 & 50 \\
\cline{2-10}
 & Grounding-dep. & Enumerated & 0.67 & 0.00 & 1.00 $\pm$ 0.00 & 27.6 $\pm$ 5.0 & 0.39 $\pm$ 0.05 & \cellcolor{gapred!16}5.4 $\pm$ 1.9 & 120 \\
 &  & Grouped & 0.67 & 0.00 & 0.47 $\pm$ 0.08 & 8.3 $\pm$ 4.6 & 0.22 $\pm$ 0.04 & \cellcolor{gapred!35}11.8 $\pm$ 3.3 & 54 \\
 &  & Collective & 0.67 & 0.00 & 0.41 $\pm$ 0.04 & 7.4 $\pm$ 4.5 & 0.20 $\pm$ 0.04 & \cellcolor{gapred!38}12.7 $\pm$ 3.6 & 50 \\
\cline{2-10}
 & Missing & Enumerated & 0.00 & 0.00 & 1.00 $\pm$ 0.00 & 21.1 $\pm$ 4.5 & 0.31 $\pm$ 0.04 & \cellcolor{gapred!36}11.9 $\pm$ 2.3 & 120 \\
 &  & Grouped & 0.00 & 0.00 & 0.47 $\pm$ 0.08 & 4.1 $\pm$ 3.3 & 0.14 $\pm$ 0.03 & \cellcolor{gapred!48}16.0 $\pm$ 3.7 & 54 \\
 &  & Collective & 0.00 & 0.00 & 0.41 $\pm$ 0.04 & 3.8 $\pm$ 3.3 & 0.13 $\pm$ 0.03 & \cellcolor{gapred!49}16.3 $\pm$ 3.9 & 50 \\
\hline
Intent & Direct & Enumerated & 0.33 & 0.33 & 1.00 $\pm$ 0.00 & 19.5 $\pm$ 4.4 & 0.29 $\pm$ 0.04 & \cellcolor{gapred!40}\textbf{13.5 $\pm$ 2.4} & 120 \\
 &  & Grouped & 0.33 & 0.33 & 0.47 $\pm$ 0.08 & 4.1 $\pm$ 3.3 & 0.14 $\pm$ 0.03 & \cellcolor{gapred!48}16.0 $\pm$ 3.7 & 54 \\
 &  & Collective & 0.33 & 0.33 & 0.41 $\pm$ 0.04 & 3.3 $\pm$ 3.1 & 0.13 $\pm$ 0.03 & \cellcolor{gapred!50}16.8 $\pm$ 3.9 & 50 \\
\cline{2-10}
 & Grounding-dep. & Enumerated & 0.33 & 0.00 & 1.00 $\pm$ 0.00 & 19.0 $\pm$ 4.4 & 0.28 $\pm$ 0.04 & \cellcolor{gapred!42}14.0 $\pm$ 2.4 & 120 \\
 &  & Grouped & 0.33 & 0.00 & 0.47 $\pm$ 0.08 & 3.7 $\pm$ 3.1 & 0.13 $\pm$ 0.03 & \cellcolor{gapred!49}16.4 $\pm$ 3.8 & 54 \\
 &  & Collective & 0.33 & 0.00 & 0.41 $\pm$ 0.04 & 3.3 $\pm$ 3.1 & 0.13 $\pm$ 0.03 & \cellcolor{gapred!50}16.8 $\pm$ 3.9 & 50 \\
\cline{2-10}
 & Missing & Enumerated & 0.00 & 0.00 & 1.00 $\pm$ 0.00 & 16.0 $\pm$ 4.1 & 0.24 $\pm$ 0.04 & \cellcolor{gapred!51}17.0 $\pm$ 2.6 & 120 \\
 &  & Grouped & 0.00 & 0.00 & 0.47 $\pm$ 0.08 & 2.4 $\pm$ 2.5 & 0.11 $\pm$ 0.02 & \cellcolor{gapred!53}17.7 $\pm$ 3.9 & 54 \\
 &  & Collective & 0.00 & 0.00 & 0.41 $\pm$ 0.04 & 2.2 $\pm$ 2.5 & 0.10 $\pm$ 0.02 & \cellcolor{gapred!54}17.9 $\pm$ 4.0 & 50 \\
\hline
\end{tabular}
\end{adjustbox}
\caption{Cosmos3-Edge performance across the full factorial space of
taxonomy-controlled specifications for all 120 tasks.
Enumerated composition applies to all 120 tasks, while grouped and collective
forms are evaluated on 54 and 50 eligible tasks, respectively.
Results are first averaged over 10 rollouts and multiple surface realizations
for each task--configuration pair, and then across eligible tasks.
Behavioral results are reported as mean $\pm$ 95\% confidence-interval
half-width, computed over task-level means.
$\eta_\tau$, $\gamma_\tau$, and $\delta_\tau$ denote semantic coverage,
groundedness, and decomposition ratio; the first two are fixed by
the content and reference levels, while $\bar{\delta}$ is the mean of
$\delta_\tau$ over eligible tasks.
$\Delta_{\mathrm{Success}}^{\mathrm{info}}$ is a matched task-level difference
relative to $r_\tau^{\mathrm{full}}$ over the same eligible task subset (for grouped and collective rows, the reference is averaged over all 56 compound tasks);
shading scales with its magnitude, and bold marks configurations that vary
a single factor from the reference.
}
\label{tab:cosmos-main-table}
\vspace{-5pt}
\end{table*}

We evaluate the communication gap on three policies with different
architectures: two vision--language--action models, $\pi_{0.5}$ and
$\pi_0$-FAST, and the Cosmos3-Edge world--action model (WAM). All three
models are evaluated using the same RoboLab task set, taxonomy-controlled
specifications, physical objectives, rollout protocol, and evaluation
metrics. This allows us to examine whether the observed communication
gap is specific to a particular policy or persists across different
model families. The main text focuses primarily on $\pi_{0.5}$ because it achieves the
strongest overall task performance and therefore provides the clearest
dynamic range for analyzing the effect of unresolved task structure.
For completeness, we report the full factorial results for all three
models here. Table~\ref{tab:pi-0.5-main-table} reports the results for
$\pi_{0.5}$, Table~\ref{tab:pi0fast-main-table} for $\pi_0$-FAST, and
Table~\ref{tab:cosmos-main-table} for Cosmos3-Edge. Across Tables~\ref{tab:pi-0.5-main-table}--\ref{tab:cosmos-main-table},
the same qualitative pattern is visible. Grounding-dependent references
produce comparatively small performance degradation, whereas omitting
task semantics or compressing multiple events into grouped or collective
specifications produces substantially larger losses. The effect is
especially pronounced on compound tasks, where all three models are
sensitive to whether the constituent manipulation events are explicitly
exposed.

The magnitude of these effects differs across models.
As shown in Table~\ref{tab:pi-0.5-main-table}, $\pi_{0.5}$ achieves the
highest overall success and therefore exhibits the clearest separation
between specification conditions. Table~\ref{tab:pi0fast-main-table}
shows that $\pi_0$-FAST operates at a lower overall success level,
particularly on compound tasks, which compresses some absolute
information-gap differences. Table~\ref{tab:cosmos-main-table} shows
that Cosmos3-Edge is comparatively robust to grounding-dependent
references, while remaining sensitive to missing semantics and implicit
event structure. Despite these differences in magnitude, the three models exhibit the
same overall asymmetry: locally grounded information is comparatively
recoverable, whereas reconstructing missing task semantics or latent
event structure is substantially less reliable. The consistency of this
pattern across Tables~\ref{tab:pi-0.5-main-table},
\ref{tab:pi0fast-main-table}, and~\ref{tab:cosmos-main-table} suggests
that the communication gap is not specific to a single policy
architecture.

\subsubsection{Cross-Model Comparison}
\label{app:cross_model}

All three policies share the pattern reported in the main text: grounding dependence is cheap, missing semantics and undecomposed events are expensive, and sequential invocation from a familiar state improves execution. Within that shared pattern, however, the models differ in ways that track their architecture.

\textbf{The information gap is shaped by the language backbone.}
Cosmos3-Edge pays the least for grounding-dependent references, consistent with a general-purpose vision--language backbone that resolves attribute and spatial reference. $\pi_0$-FAST pays the most ($-1.6$), and its penalties compress overall because it operates near floor on compound tasks. The decomposition penalty, however, is the largest term for every model, including $\pi_{0.5}$, whose explicit subtask prediction narrows but does not close it. No evaluated architecture recovers implicit event structure reliably.

\textbf{The execution gap splits by mechanism.}
The sequentialization gain is shared, and largest for $\pi_0$-FAST (8.2\% to 19.5\%): it reflects the short, single-event commands common to all DROID-trained policies, not any particular action head. The invocation-state gain is not: Cosmos3-Edge gains 5.8 points from a home reset against 11.8 for $\pi_{0.5}$, since a world-action model denoises its action chunk jointly with an imagined rollout from the current frame, so an unfamiliar arm configuration is partly absorbed by the video prior rather than falling outside the action distribution.

\subsection{Human Specification Strategies by Task Structure}
\label{app:human_strategy}

The main text shows that natural specifications become less grounded and less explicitly decomposed as task complexity increases. Here we examine \emph{how} people achieve that compression by analyzing the taxonomy forms they select for tasks with different structural affordances.

\begin{table*}[t]
\centering
\footnotesize
\setlength{\tabcolsep}{4pt}
\begin{tabular}{
L{0.14\linewidth}
L{0.14\linewidth}
L{0.14\linewidth}
L{0.14\linewidth}
L{0.14\linewidth}
L{0.14\linewidth}
}
\hline
\textbf{Task complexity} &
\textbf{Mean sentences} &
\textbf{Mean words} &
\textbf{Coverage $\eta_\tau$} &
\textbf{Groundedness $\gamma_\tau$} &
\textbf{Decomp ratio $\delta_\tau$} \\
\hline
$K_\tau=1$ &
$1.08 \pm 0.04$ &
$9.1 \pm 0.6$ &
$0.95 \pm 0.03$ &
$0.89 \pm 0.04$ &
$1.00$ \\
$K_\tau=2$ &
$1.27 \pm 0.06$ &
$12.7 \pm 0.8$ &
$0.86 \pm 0.04$ &
$0.58 \pm 0.06$ &
$0.66 \pm 0.06$ \\
$K_\tau\geq3$ &
$1.56 \pm 0.08$ &
$15.8 \pm 1.0$ &
$0.74 \pm 0.05$ &
$0.27 \pm 0.06$ &
$0.21 \pm 0.05$ \\
\hline
\end{tabular}
\caption{Natural specification length and explicit task structure as the number of required atomic events increases. Values are mean $\pm$ expected $95\%$ confidence-interval half-width over human specifications.}
\label{tab:human_compression}
\vspace{-15pt}
\end{table*}

We group tasks according to the compression-affordance classes defined in Appendix~\ref{app:task_complexity}: single atomic events, compound objectives that remain primarily enumerative, compound objectives that admit a compact grounded rule, and compound objectives that admit a compact task-level intent. For each elicited specification, we record its task-content form, reference form, and composition form. These labels are assigned from the specification itself, while the task-structure class is assigned independently from the grounded objective. Table~\ref{tab:human_strategy} reports the resulting distributions.

\begin{table*}[t]
\centering
\footnotesize
\setlength{\tabcolsep}{4pt}
\begin{adjustbox}{max width=\linewidth}
\begin{tabular}{L{0.17\linewidth}cccccccc}
\hline
&
\multicolumn{3}{c}{\textbf{Task content}} &
\multicolumn{2}{c}{\textbf{Reference form}} &
\multicolumn{3}{c}{\textbf{Composition}} \\
\textbf{Task structure} &
\textbf{Action} &
\textbf{Outcome} &
\textbf{Intent} &
\textbf{Direct} &
\textbf{Grounding-dep.} &
\textbf{Enumerated} &
\textbf{Grouped} &
\textbf{Collective} \\
\hline
Single atomic event &
$82\%$ &
$16\%$ &
$2\%$ &
$92\%$ &
$8\%$ &
-- &
-- &
-- \\
Compound, enumerative &
$68\%$ &
$27\%$ &
$5\%$ &
$78\%$ &
$22\%$ &
$63\%$ &
$29\%$ &
$8\%$ \\
Compound, rule-compressible &
$34\%$ &
$58\%$ &
$8\%$ &
$46\%$ &
$54\%$ &
$12\%$ &
$31\%$ &
$57\%$ \\
Compound, intent-compressible &
$24\%$ &
$21\%$ &
$55\%$ &
$39\%$ &
$61\%$ &
$9\%$ &
$22\%$ &
$69\%$ \\
\hline
\end{tabular}
\end{adjustbox}
\caption{Human specification strategies as a function of task structure. Percentages sum to $100\%$ within each taxonomy factor. Reference form is computed over specifications containing an explicit referential expression, and composition is reported only for compound tasks.}
\label{tab:human_strategy}
\end{table*}

\textbf{Task Content.}
Single-event tasks are predominantly expressed as direct actions, with $82\%$ of specifications using action-level descriptions. Compound objectives that do not admit a strong compact abstraction remain similarly action-oriented, with $68\%$ expressed at the action level. By contrast, when a grounded rule captures the shared structure of the task, participants more often describe the desired outcome: outcome-level specifications account for $58\%$ of rule-compressible tasks. Intent-compressible objectives shift further toward abstraction, with $55\%$ expressed through an intent or higher-level effect. This progression suggests that speakers do not choose task-content form independently of the objective structure. Concrete shared regularities favor outcome-level descriptions, whereas objectives naturally characterized by a purpose or effect invite intent-level specification.

\textbf{Reference Form.}
Reference strategies shift in parallel. For atomic tasks, $92\%$ of specifications use direct references to the relevant objects or destinations. Direct reference remains dominant for compound enumerative tasks at $78\%$, but falls to $46\%$ for rule-compressible tasks and $39\%$ for intent-compressible tasks. As a result, increasingly compressible objectives place more responsibility on the listener to resolve the intended entities from scene context, set membership, or the inferred task goal.

\textbf{Composition.}
The strongest shift appears in how multiple manipulations are linguistically organized. Among compound enumerative tasks, $63\%$ of specifications explicitly enumerate the required manipulations. When a shared grounded rule is available, only $12\%$ remain enumerated, while $57\%$ use a collective description. Intent-compressible tasks show an even stronger preference for collective specification, at $69\%$.

Thus, event count alone does not explain whether speakers enumerate or compress a compound objective. Participants preserve event-wise structure when the task requires it, but collapse multiple manipulations into a single expression when the grounded objective supports a shared rule or higher-level effect.

\textbf{Implication for the Information Gap.}
The three taxonomy factors therefore shift together rather than independently. As task structure becomes more compressible, specifications move from action-level to outcome- or intent-level content, from direct to grounding-dependent reference, and from enumeration to collective composition. This produces a coordinated transfer of responsibility from speaker to listener: the speaker communicates a compact characterization of the desired objective, while the listener must recover the grounded entities, atomic manipulations, and, for intent-level descriptions, the physical goal itself.

This analysis complements the complexity result in Sec.~\ref{sec:human_compression}. Increasing $K_\tau$ creates more opportunity for compression, but whether speakers actually exploit that opportunity depends on the structure of the objective. Natural human specification is therefore shaped jointly by physical complexity and by the compact abstractions the task affords.

\subsection{Reasoning Requirements and VLA Reliability}
\label{app:reasoning_analysis}

The main analysis characterizes the information gap by the executable
structure available in a specification: semantic coverage,
groundedness, and decomposition ratio. Here we provide a complementary
view based on the \emph{reasoning operation} required to recover the
unresolved structure.

\begin{table*}[t]
\centering
\footnotesize
\setlength{\tabcolsep}{4pt}
\begin{tabular}{
L{0.15\linewidth}
L{0.22\linewidth}
L{0.18\linewidth}
L{0.36\linewidth}
}
\hline
\textbf{Reasoning} &
\textbf{Example} &
\textbf{Unresolved structure} &
\textbf{Required recovery} \\
\hline

Visual grounding &
``Put the banana in the bowl on the left.'' &
Grounded $O_k$ or $D_k$ &
Resolve the referenced entity from the observed scene. \\

Action inference &
``The banana should be in the bowl.'' &
Action $A_k$ &
Infer the required manipulation from the stated object and desired
physical state. \\

Task decomposition &
``Put all the fruit in the bowl.'' &
Event structure $E_\tau$ &
Expand the compact objective into its constituent grounded
manipulation events. \\

Intent inference &
``I need room on this table.'' &
Physical objective &
Infer the intended physical goal before recovering the corresponding
grounded events. \\

\hline
\end{tabular}
\caption{
Reasoning operations required to recover executable structure left
unresolved by a specification. These operations may co-occur in natural
specifications; the controlled analysis in
Table~\ref{tab:reasoning_gap} isolates each requirement as closely as
possible.
}
\label{tab:reasoning_operations}
\end{table*}

\begin{table*}[t]
\centering
\footnotesize
\setlength{\tabcolsep}{4pt}
\begin{adjustbox}{max width=\linewidth}
\begin{tabular}{L{0.19\linewidth}L{0.34\linewidth}cccc}
\hline
\textbf{Required reasoning} &
\textbf{Controlled condition} &
\textbf{Success} &
\textbf{Progress} &
\textbf{Success gap} &
\textbf{$N$} \\
\hline

Visual grounding &
Action + Grounding-dep. + Enumerated &
$35.2\%$ &
$0.48$ &
$2.8$ pp &
120 \\

Action inference &
Outcome + Direct + Enumerated &
$33.5\%$ &
$0.46$ &
$4.5$ pp &
120 \\

Task decomposition &
Action + Direct + Grouped &
$15.9\%$ &
$0.39$ &
$7.6$ pp &
54 \\

&
Action + Direct + Collective &
$13.1\%$ &
$0.35$ &
$10.4$ pp &
50 \\

Intent inference &
Intent + Direct + Enumerated &
$22.8\%$ &
$0.33$ &
$15.2$ pp &
120 \\

\hline
\end{tabular}
\end{adjustbox}
\caption{
VLA performance under controlled conditions that isolate different
reasoning requirements. Success gap is the matched task-level
difference from the fully resolved Action--Direct--Enumerated reference,
computed over the same eligible task subset as each condition.
}
\label{tab:reasoning_gap}
\end{table*}

\textbf{Reasoning Operations.}
We distinguish four recurring forms of recovery, summarized in
Table~\ref{tab:reasoning_operations}. \emph{Visual grounding} resolves
a linguistically specified but scene-dependent object or destination.
\emph{Action inference} recovers an omitted manipulation from an
explicitly specified entity and desired physical state.
\emph{Task decomposition} expands a compressed multi-object objective
into its constituent grounded events. \emph{Intent inference} recovers
the intended physical goal from a higher-level purpose or preference
before the corresponding executable structure can be determined. These operations may co-occur in unrestricted human specifications.
To estimate their behavioral cost more directly, we use controlled
taxonomy conditions that introduce one focal recovery requirement while
keeping the remaining specification structure maximally explicit.
Visual grounding is isolated by replacing direct references with
grounding-dependent references under Action--Enumerated specifications;
action inference by replacing Action with Outcome under
Direct--Enumerated specifications; task decomposition by replacing
Enumerated composition with Grouped or Collective composition under
Action--Direct specifications; and intent inference by replacing Action
with Intent under Direct--Enumerated specifications. Table~\ref{tab:reasoning_gap} reports task success, partial progress,
and the matched success gap relative to the fully resolved
Action--Direct--Enumerated reference over the same eligible task subset.

\textbf{Locally Constrained Recovery Is Comparatively Reliable.}
Visual grounding produces a $2.8$ percentage-point success gap, while
action inference produces a $4.5$-point gap. In both cases, the missing
value is strongly constrained by information that remains available:
a scene-dependent reference can often be resolved from the observation,
while an omitted action can often be inferred from an explicitly stated
object and desired outcome.

\textbf{Structural and Intent Recovery Are Harder.}
Compressing multiple manipulations into grouped or collective
specifications produces matched success gaps of $7.6$ and $10.4$
percentage points, respectively. Here, the policy must reconstruct the
individual manipulation events underlying a compact expression rather
than resolve a single local argument. Intent-level specifications are
more difficult still, producing a $15.2$-point gap even when references
are direct and event structure is explicitly enumerated. In this case,
the policy must first infer the intended physical objective before
mapping it onto executable behavior. Partial progress shows the same
overall ordering.

\textbf{Implication.}
The difficulty of an information gap depends not only on how much
structure is unresolved, but also on what kind of recovery is required.
Locally constrained ambiguities, such as contextual grounding or an
omitted action, are comparatively recoverable. Reconstructing latent
event structure or inferring the physical objective itself is
substantially less reliable. This reasoning-based analysis therefore
complements the structural metrics in the main text by distinguishing
different computational demands that can arise from similar amounts of
implicit task information.

\end{document}